\pdfoutput=1
\documentclass[sigconf,]{acmart}
\copyrightyear{2018} 
\acmYear{2018} 
\setcopyright{acmcopyright}
\acmConference[KDD '18]{The 24th ACM SIGKDD International Conference on Knowledge Discovery \& Data Mining}{August 19--23, 2018}{London, United Kingdom}
\acmPrice{15.00}
\acmDOI{10.1145/3219819.3220021}
\acmISBN{978-1-4503-5552-0/18/08}

\usepackage{booktabs} 

\usepackage{subcaption,verbatim,engord,enumitem,multirow,bm,caption}
\usepackage{algorithm,algorithmic}
\usepackage{amsmath,dsfont}
\usepackage{amssymb}
\usepackage{graphicx}
\usepackage{amscd}
\usepackage{mathtools}
\usepackage{balance}

\DeclarePairedDelimiter{\floor}{\lfloor}{\rfloor}
\DeclareMathOperator*{\argmin}{argmin}

\begin{document}
\title{Ranking Distillation: Learning Compact Ranking Models With High Performance for Recommender System}

\author{Jiaxi Tang}
\affiliation{%
  \institution{School of Computing Science\\Simon Fraser University}
  \city{British Columbia} 
  \state{Canada} 
}
\email{jiaxit@sfu.ca}

\author{Ke Wang}
\affiliation{%
  \institution{School of Computing Science\\Simon Fraser University}
  \city{British Columbia} 
  \state{Canada} 
}
\email{wangk@cs.sfu.ca}

\fancyhead{} 

\begin{abstract}

We propose a novel way to train ranking models, such as recommender systems, that are both effective and efficient. Knowledge distillation (KD) was shown to be successful in image recognition
to achieve both effectiveness and efficiency. We propose a KD technique
for learning to rank problems, called \emph{ranking distillation (RD)}. Specifically, we train a smaller student model to
learn to rank documents/items from both the training data and the supervision of a larger teacher model.
The student model achieves a similar ranking performance to that of the large teacher model,
but its smaller model size makes the online inference more efficient.
RD is flexible because 
it is orthogonal to the choices of ranking models for the teacher and student.  
We address the challenges of RD for ranking problems. 
The experiments on public data sets and state-of-the-art recommendation models showed that
RD achieves its design purposes: the student model learnt with RD has a model size less than half of the teacher model while achieving a ranking performance similar to
the teacher model and much better than the student model learnt without RD.

\end{abstract}



\maketitle

\section{introduction}\label{sec:intro}
In recent years, information retrieval~(IR) systems become a core technology of many large companies, such as web page retrieval for Google and Yahoo; personalized item retrieval \emph{a.k.a} recommender systems for  Amazon and Netflix. The core in such systems is a ranking model for computing the relevance score of each ($q$, $d$) pair for future use, where $q$ is the query (\emph{e.g.,} keywords for web page retrieval and user profile for recommender systems) and $d$ is a document (\emph{e.g.,} a web page or item). The \emph{effectiveness} of IR systems largely depends on how well the ranking model performs, whereas the \emph{efficiency} determines how fast the systems will respond to user queries, \emph{a.k.a} online inferences.


Since the 2009 Netflix Prize competition, it is increasingly realized that a simple linear model with few parameters cannot model the complex query-document (user-item) interaction properly, and latent factor models with numerous
parameters are shown to have better effectiveness and good representation power~\cite{koren2009matrix}.
Recently, with the great impact of neural networks on computer vision~\cite{krizhevsky2012imagenet, karpathy2014large} and natural language processing~\cite{mikolov2010recurrent, yoon14convolution}, a new branch of IR using neural networks 
has shown strong performances.
As neural networks have incredible power to automatically capture features, recent works use neural networks to capture semantic representations for both queries and documents, relieving the manual feature engineering work required by other approaches. 
Several successful ranking models with neural networks have been investigated
~\cite{he2017neural, tang2018caser, xiong2017ena, wang2015collaborative}. However, 
 the size of such models~(in terms of the number of model parameters)
increases by an order of magnitude or more than previous methods.
While such models have better ranking performance by capturing more query-document interactions,
they incur a larger latency at online inference phase when responding to user requests due to the larger model size.

Balancing effectiveness and efficiency has been a line of recent research~\cite{zhang2016discrete, zhou2012learning, zhang2014preference, teflioudi2015lemp, li2017fexipro}. Discrete hashing techniques~\cite{zhou2012learning, zhang2016discrete, zhang2017discrete} and binary coding of model parameters are suggested to speed up the calculation of the relevance score for a given ($q$,$d$) pair.  Other works focus on database-related methods, such as pruning and indexing to speed-up retrieval of related items~\cite{teflioudi2015lemp,li2017fexipro}, using fast models for candidate generation and applying time-consuming models to the candidates for online inferences~\cite{covington2016deep,liu2017related}.
These methods either lose much of effectiveness, due to the introduced model constraints, or cannot be easily extended to other models in most cases, due to the model-dependency nature.

\begin{figure}[t!]  
        \centering
        \begin{subfigure}[b]{0.35\textwidth}
                \includegraphics[width=\textwidth]{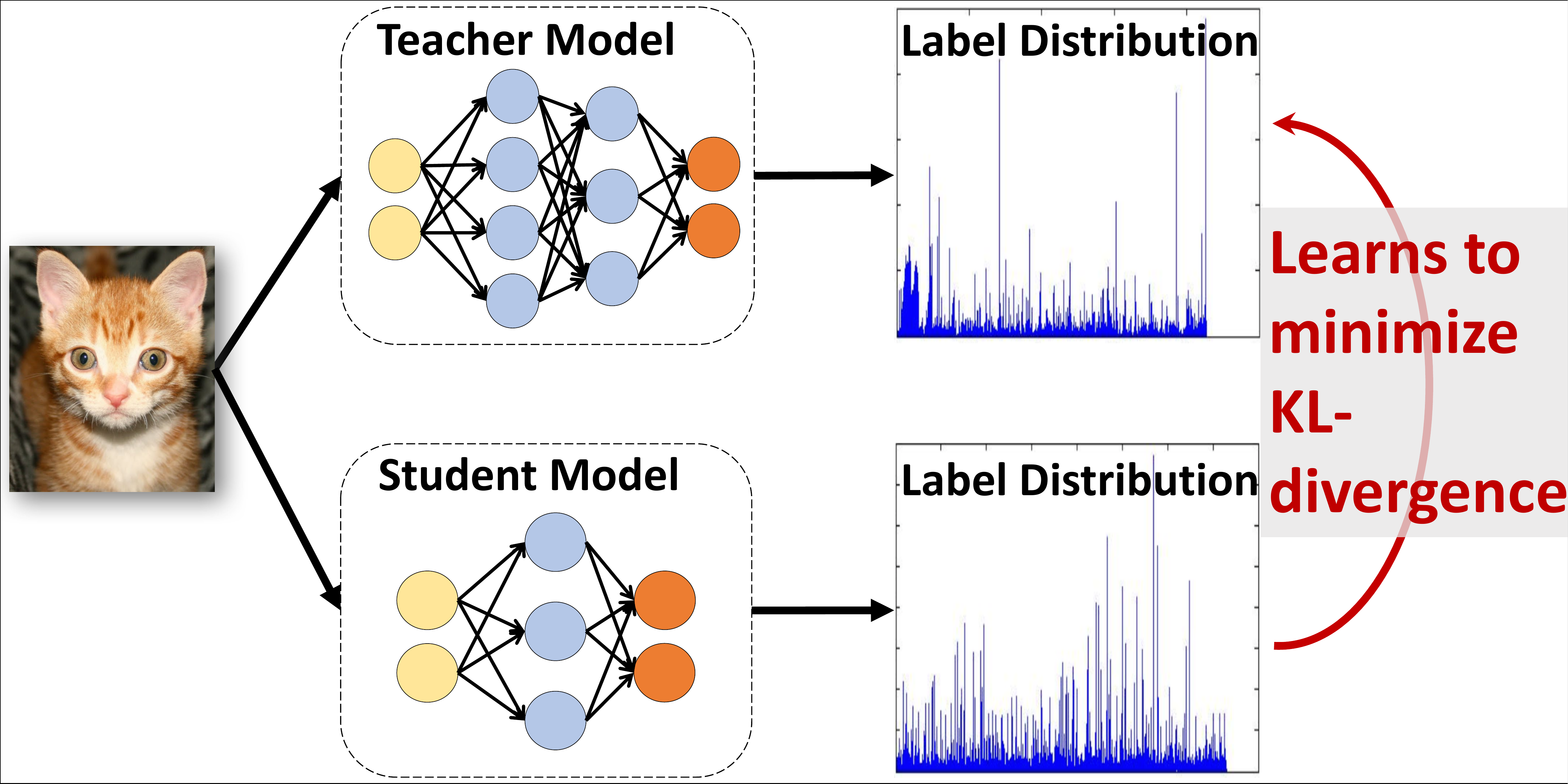}
                \caption{Knowledge Distillation for classification}
                \label{fig:kd_rd1}
        \end{subfigure}%
        \vspace{0.20cm}
        \begin{subfigure}[b]{0.35\textwidth}                                 \includegraphics[width=\textwidth]{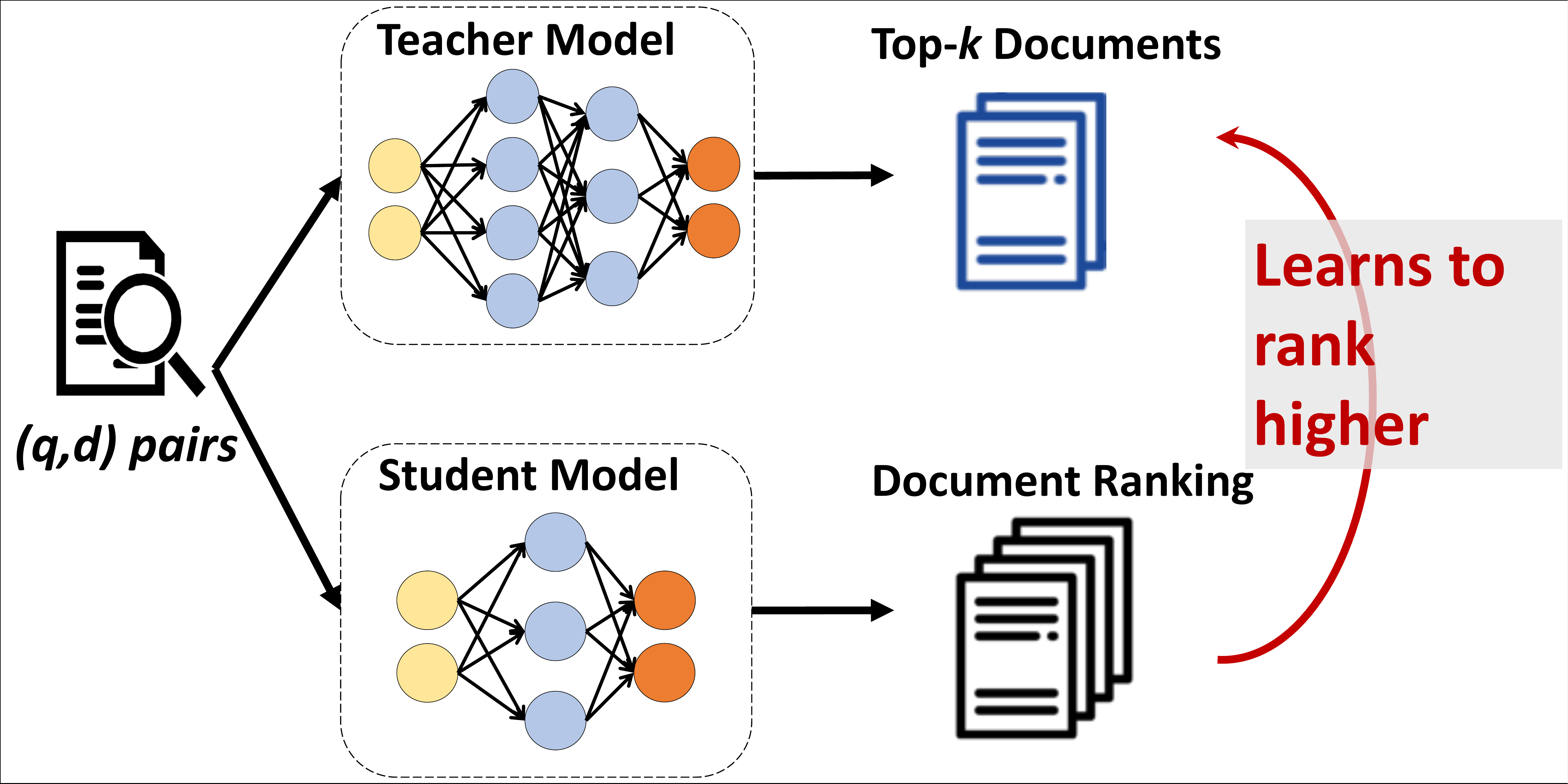}
                \caption{Ranking
                Distillation for ranking problems}
                \label{fig:kd_rd2}
        \end{subfigure}
        \vspace{0.20cm}
        \caption{(a) Knowledge Distillation: given an input, student model learns to minimize the KL Divergence of its label distribution and teacher model's. (b) Ranking Distillation: given an query, student model learns to give higher rank for it's teacher model's top-$K$ ranking of documents.}\label{fig:kd_rd}
\end{figure}

\subsection{Knowledge Distillation}\label{sec:limitations}

\emph{Knowledge distillation} (KD), a.k.a., knowledge transfer, is a model-independent strategy for generating a compact model for better inference efficiency while retaining the model effectiveness~\cite{hinton2015distilling, ba2014deep, Anil2018Large}. The idea of KD is shown in Figure~\ref{fig:kd_rd1} for image recognition. During
the offline training phase, a larger \emph{teacher model} is first trained from the training set, and
a smaller \emph{student model} is then trained by minimizing two deviations: the deviation from the
training set's ground-truth label distribution, and the deviation from the label distribution generated
by the teacher model. Then the student model is used for making online inferences. Intuitively,
the larger teacher model helps capture more information of the label distribution (for example, outputting a high probability for ``tiger" images for a ``cat" image query due to correlation), which is used
as an additional supervision to the training of the student model. The student model
trained with KD has an effectiveness comparable to that of the teacher model~\cite{hinton2015distilling, kim2016sequence, Anil2018Large} 
and can make more efficient online inference due to its small model size.

Despite of this breakthrough in image recognition, it is not straightforward to apply KD to ranking models and ranking problems~(\emph{e.g.,} recommendation). First, the existing KD is designed for classification problems, not for ranking problems. In ranking problems, the focus is on predicting the relative order of documents or items, instead of 
predicting a label or class as in classification. Second, KD requires computing the label distribution of documents for each query using both teacher and student models, which is feasible for image classification where there are a small number of labels, for example, no more than 1000 for the ImageNet data set; however, for ranking and recommendation problems, the total number of documents or items could be several orders of magnitudes larger, say millions, and
computing the distribution for all documents or items for each training instance makes little sense, especially only the highly ranked 
documents or items near the top of the ranking will matter. We also want to point out that, for context sensitive recommendation, such as sequential recommendation, the items to be recommended usually depend on the user behaviors prior to the recommendation point (e.g., what she has viewed or purchased), and the set of contexts of a user is only known at the recommendation time. This feature requires recommender system to be strictly responsive and makes online inference efficiency particularly important.

\subsection{Contributions}
In this work, we study knowledge distillation for the learning to rank problem that is the core in recommender systems and many other IR systems. Our objective is achieving the ranking performance of
a large model with the online inference efficiency of a small model. In particular, by fusing the idea of knowledge transfer and learning to rank, we propose a technique called
\emph{ranking distillation} (RD) to learn a compact ranking model that remains effective. 
The idea is shown in Figure~\ref{fig:kd_rd2} where a small student model is trained to learn to rank from two sources of information, \emph{i.e.}, the training set and the top-$K$ documents for each query generated by a large well-trained teacher ranking model. 
With the large model size, the teacher model captures more ranking patterns from the training set
and provides top-$K$ ranked unlabeled documents as an extra training data for the student model. This makes RD differ from KD, as teacher model in KD only generate additional labels on existing data, while RD generate additional training data and labels from unlabeled data set.
The student model benefits from the extra training data
generated from the teacher, in addition to the data from usual training set, thus, 
inherits the strong ranking performance of the teacher,
but is more efficient for online inferences thanks to its small model size.


We will examine several key issues of RD, \emph{i.e.}, the problem formulation, 
the representation of teacher's supervision, and the balance between the trust 
on the data from the training set and the data generated by the teacher model, and present our solutions.
Extensive experiments on recommendation problems and real-world datasets show that the student model achieves
a similar or better ranking performance compared to the teacher model while using less than half of model parameters. While the design goal of RD is retaining the teacher's effectiveness (while achieving the student's online inference efficiency), RD exceeds this expectation that the student sometime has
a even better ranking performance than the teacher. Similar to KD,
RD is orthogonal to the choices of student and teacher models by treating them as black-boxes. 
To our knowledge, this is the first attempt to adopt the idea of knowledge distillation to large-scale ranking problems.

%
%

In the rest of the paper, we introduce background materials in Section \ref{sec:pre}, propose ranking distillation for ranking problems in Section~\ref{sec:model}, present experimental studies in Section~\ref{sec:exper}, and discuss
related work in Section~\ref{sec:related}. We finally conclude with a summary of this work and in Section~\ref{sec:conclusion}.

\section{Backgrounds}\label{sec:pre}
We first review the learning to rank problem, then revisit the issues of effectiveness and efficiency in the problem, which serves to motivate our ranking distillation. Without loss of generality, we use the IR terms ``query" $q$ and ``document" $d$ in our discussion, but these terms can be replaced with ``user profile" and ``item" when applied to recommender systems.

\subsection{Ranking from scratch}
The learning to rank problem can be summarized as follows: Given a set of queries $\mathcal{Q}$=\{$q_1$,$\cdots$,$q_{|\mathcal{Q}|}$\} and a set of documents $\mathcal{D}$=\{$d_1$,$\cdots$,$d_{|\mathcal{D}|}$\}, we want to retrieve documents
that are most relevant to a certain query.
The degree of relevance for a query-document pair $(q,d)$
is determined by a relevance score. Sometimes, for a single $(q,d)$ pair,
a relevance score $y^{(q)}_d$ is labeled by human (or statistical results) as ground-truth, but the number of labeled $(q,d)$ pairs is much smaller compared to the pairs with unknown labels. Such labels can be binary~(\emph{i.e.,} relevant/non-relevant) or ordinal~(\emph{i.e.,} very relevant/relevant/non-relevant). In order to rank documents for future queries with unknown relevance scores, we need a ranking model to predict their relevance scores.
A ranking model $M(q,d;\boldsymbol{\theta})=\hat{y}^{(q)}_d$ is defined by a set of model parameters $\boldsymbol{\theta}$
and computes a relevance score $\hat{y}^{(q)}_d$ given the query $q$ and document $d$. The model
predicted document ranking is supervised by the human-labeled ground truth ranking.
The optimal model parameter set $\boldsymbol{\theta}^{*}$ 
is obtained by minimizing a ranking-based loss function:
\begin{equation}\label{eqn:rankloss_general}
\boldsymbol{\theta}^{*} = \argmin_{\boldsymbol{\theta}}\sum_{q \in \mathcal{Q}}\mathcal{L}^{R}(\boldsymbol{y}^{(q)}, \boldsymbol{\hat{y}}^{(q)}).
\end{equation}
For simplicity, we focus on a single query $q$ and omit the superscripts related to queries (\emph{i.e.,} $y_d^{(q)}$ will becomes $y_d$).

The ranking-based loss could be categorized as point-wise, pair-wise, and list-wise. Since the first two are more widely adopted, we don't discuss list-wise loss in this work.
%
The point-wise loss is widely used when relevance labels are binary~\cite{he2017neural,tang2018caser}. 
One typical point-wise loss is taking the negative logarithmic of the likelihood function:
\begin{equation}\label{eqn:rankloss_point}
\begin{aligned}
\mathcal{L}^{R}(\boldsymbol{y}, \boldsymbol{\hat{y}})=-(&\sum_{d\in \boldsymbol{y}_{d+}}\text{log}(P(rel=1|\hat{y}_d)) \\ + &\sum_{d\in \boldsymbol{y}_{d-}}\text{log}(1 - P(rel=1|\hat{y}_d))),
\end{aligned}
\end{equation}
where $\boldsymbol{y}_{d+}$ and $\boldsymbol{y}_{d-}$ are the sets of relevant and non-relevant documents, respectively. We could use the \emph{logistic} function $\sigma(x)=1/(1+e^{-x})$ and $P(rel=1|\hat{y}_d)=\sigma(\hat{y}_d)$ to transform a real-valued relevance score to the probability of a document being relevant~($rel=1$). 
For ordinal relevance labels, pair-wise loss better models the partial order information:
\begin{equation}\label{eqn:rankloss_pair}
\mathcal{L}^{R}(\boldsymbol{y}, \boldsymbol{\hat{y}})=-\sum_{d_i, d_j\in \boldsymbol{C}}\text{log}(P(d_i \succ d_j|\hat{y}_i, \hat{y}_j)),
\end{equation}
where $\boldsymbol{C}$ is the set of document pairs $\{(d_i, d_j):y_i \succ y_j\}$ and the probability $P(d_i \succ d_j)$ can be modeled using the \emph{logistic} function $P(d_i \succ d_j| y_i,y_j)=\sigma(y_i-y_j)$.

\subsection{Rethinking Effectiveness and Efficiency}
We consider ranking models with latent factors or neural networks (\emph{a.k.a} neural ranking models) instead of traditional models (\emph{e.g.,} SVM, tree-based models) for the following reasons. First, these models are well-studied recently for its capability to capture features from a latent space and are shown to be highly effective; indeed, neural ranking models are powerful for capturing rich semantics for queries and documents,
which eliminates the tedious and ad-hoc feature extraction and engineering normally required in traditional models.
Second, these models usually require many parameters and suffer from efficiency issue when making online inferences.
Third, traditional models like SVM usually has convex guarantees and are trained through convex optimization.
The objectives of latent factor models and neural networks are usually non-convex~\cite{choromanska2015loss, kawaguchi2016deep}, which means that their training processes are more challenging and need more attentions.

\begin{figure}[t!]  

        \centering
        \begin{subfigure}[b]{0.23\textwidth}
                \includegraphics[width=\textwidth]{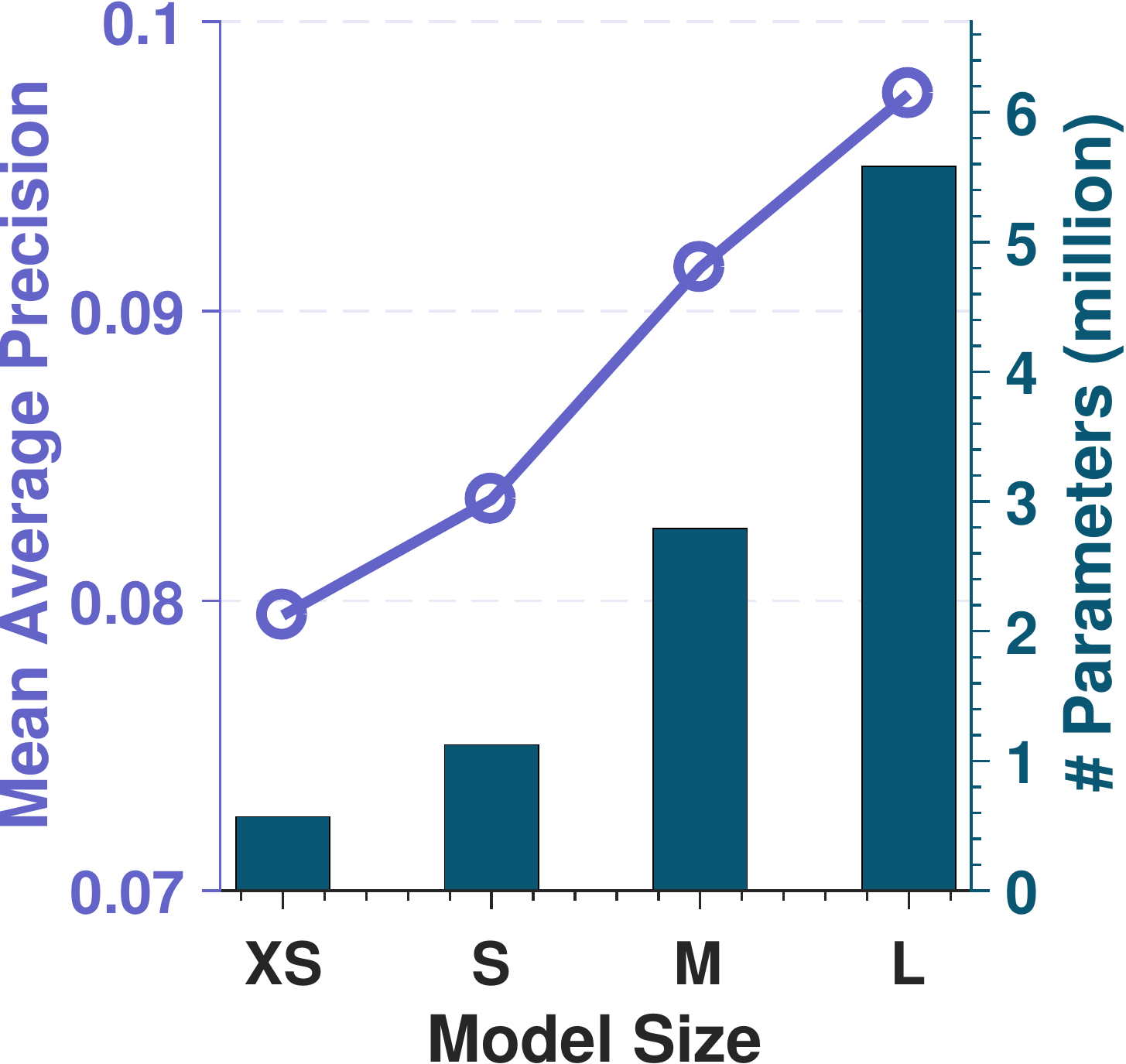}
                \caption{MAP \emph{vs.} model size}
                \label{fig:precision_dim}
        \end{subfigure}%
        ~
        \hspace{0.05in}
        \begin{subfigure}[b]{0.23\textwidth}                                     \includegraphics[width=\textwidth]{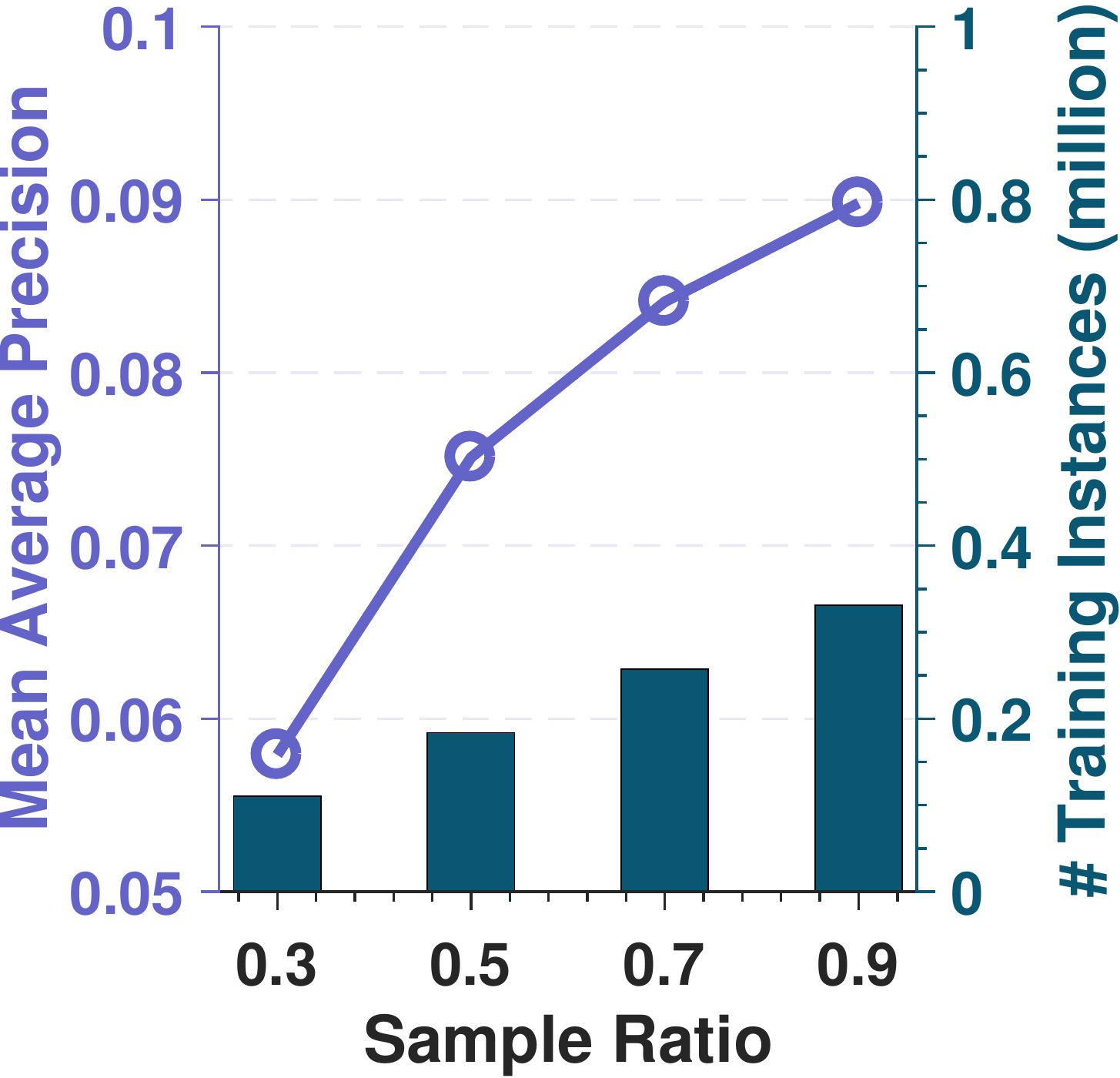}
                \caption{MAP \emph{vs.} training instances}
                \label{fig:precision_data}
        \end{subfigure}
        \caption{Two ways of boosting mean average precision~(MAP) on Gowalla data for recommendation. (a) shows that a larger model size in number of parameters, indicated by the bars, leads to a higher MAP. (b) shows that a larger sample size of training instances leads to a higher MAP.}\label{fig:precision}

\end{figure}

The goal of ranking models is predicting the rank of documents as accurately as possible 
near the top positions, through learning from human-labeled ground-truth document ranking.
Typically, there are two ways to make a ranking model perform better at top positions: (1) By having a large model size, as long as it doesn't overfit the data, the model could better capture complex query-document interaction patterns 
and has more predictive capability. Figure~\ref{fig:precision_dim} shows that, when a ranking model has more parameters, it acquires more flexibility to fit the data and has a higher MAP, where the mean average precision~(MAP) is more sensitive to the precision at top positions. (2) By having more training data, side information, human-defined rules~\emph{etc.}, the model can be trained with more guidance and has less variance in gradients~\cite{hinton2015distilling}.
Figure~\ref{fig:precision_data} shows that, when more training instances are sampled from the underlying data distribution, a ranking model could achieve a better performance. However, each method has its limitations: method (1) surrenders efficiency for effectiveness whereas method (2) requires additional informative data, which is not always available or is expensive to obtain in practice.

\section{Ranking Distillation}\label{sec:model}
In this section, we propose \emph{ranking distillation} (RD) to address the dual goals of effectiveness and efficiency for ranking problems. To address the efficiency of online inference, we use a smaller ranking model so that we can rank documents for a given query more efficiently. To address the effectiveness issue without requiring more training data, we introduce
extra information generated from a well-trained teacher model and make the student model as effective as the teacher.
%

\begin{figure}[t!]  
\centering
\includegraphics[scale=0.12]{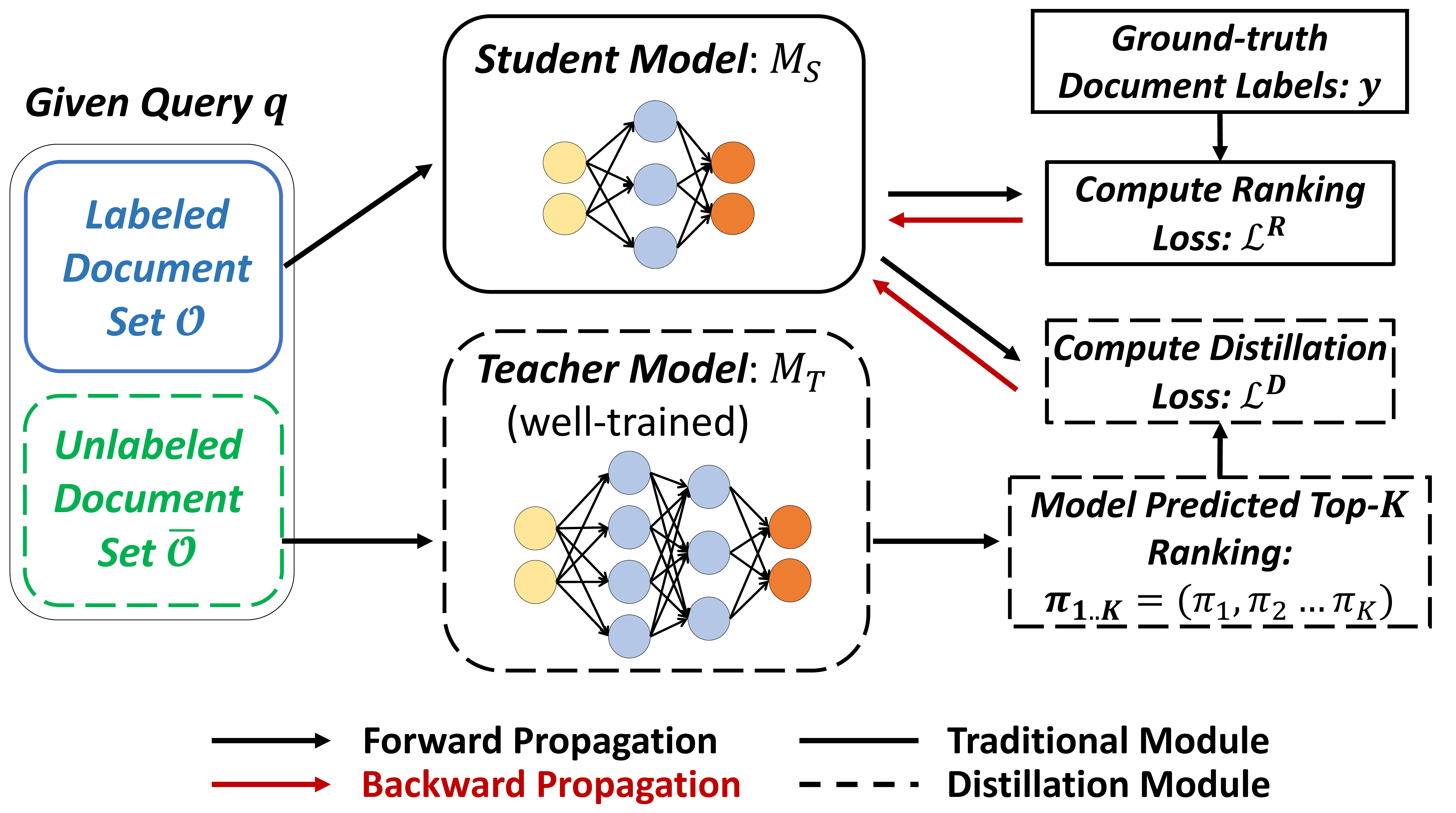}
\vspace{-0.0cm}
\caption{ The learning paradigm with ranking distillation. We first train a teacher model and let it predict a top-$K$ ranked list of unlabeled~(unobserved) documents for a given query $q$. The student model is then supervised by both ground-truth ranking from the training data set and teacher model's top-$K$ ranking on unlabeled documents. }\label{fig:distill}
\end{figure}

\subsection{Overview}\label{sec:model_overview}
Figure~\ref{fig:distill} shows the overview of ranking distillation. In the offline training phase (prior to any user query), similar to KD, first we train a large teacher model with a strong ranking performance on the training set. Then for each query, we use the well-trained teacher model to make predictions on unlabeled documents~(green part in Figure~\ref{fig:distill}) and use this extra information for learning the smaller student model. Since the teacher model is allowed to have many parameters, it captures more complex features for ranking and is much powerful, thus, its predictions on unlabeled documents could be used to provide extra information for the student model's training. The student model with fewer parameters
is more efficient for online inference, and because of the extra information
provided by the teacher model, the student model inherits the high ranking performance of the teacher model.

Specifically, the offline training for student model with ranking distillation consists of two steps. First, we train a larger teacher model $M_T$ by minimizing a ranking-based loss with the ground-truth ranking from the training data set, as showed in Eqn~(\ref{eqn:rankloss_general}).
With much more parameters in this model, it captures more patterns from data and thus has a strong performance.
We compute the predicted relevance scores of the teacher model $M_T$ for unlabeled documents $\bar{\mathcal{O}}=\{d:y_d=\emptyset\}$ and get a top-$K$ unlabeled document ranking $\boldsymbol{\pi}_{1..K}=(\pi_1,\ldots,\pi_K)$, where $\pi_r \in \mathcal{D}$ is the $r$-th document in this ranking. Then, we train a smaller ranking model $M_S$ to minimize a \emph{ranking loss} from the ground-truth ranking in the training data set, as well as a \emph{distillation loss} with the exemplary top-$K$ ranking on unlabeled document set $\boldsymbol{\pi}_{1..k}$ offered by its teacher $M_T$. The overall loss to be minimized is as follows:
\begin{equation}\label{eqn:rd1}
\mathcal{L}(\theta_S) =(1-\alpha) \mathcal{L}^{R}(\boldsymbol{y}, \boldsymbol{\hat{y}}) + \alpha \mathcal{L}^{D}(\boldsymbol{\pi}_{1..K},\boldsymbol{\hat{y}}).
\end{equation}
Here $\boldsymbol{\hat{y}}$ is the student model's predicted scores\footnote{When using point-wise and pair-wise losses, we only need to compute the student's predictions for a subset of documents, instead of all documents, for a given query.}.
$\mathcal{L}^{R}(,)$ stands for the ranking-based objective as in Eqn~(\ref{eqn:rankloss_general}). 
The distillation loss, denoted by $\mathcal{L}^{D}(,)$, uses teacher model's top-$K$ ranking on unlabeled documents to guide the student model learning. $\alpha$ is the hyper-parameter used for balancing these two losses.

For a given query, the top documents
ranked by the well-trained teacher can be regarded to have a strong correlation to this query, although they are not labeled in the training set. 
For example, if a user watches many action movies, the teacher's top-ranked documents may contain some other action movies as well as some adventure movies because they are correlated.
In this sense, the proposed RD lets the teacher model teach its student to find the correlations and capture their patterns, thus, makes the student more generalizable and perform well on unseen data in the future. We use the top-$K$ ranking from the teacher instead of the whole ranked list because the noisy ranking at lower positions tends to cause the student model to overfit its teacher and lose generalizability.
Besides, only top positions are considered important for ranking problems. $K$ is a hyper-parameter that represents the trust level on the teacher during teaching, \emph{i.e.}, how much we adopt from teacher.

The choice of the ranking loss $\mathcal{L}^{R}(,)$ follows from different models' preferences and we only focus on the second term $\mathcal{L}^{D}(,)$ in Eqn~(\ref{eqn:rd1}). A question is how much we should trust teacher's top-$K$ ranking, 
especially for a larger $K$. In the rest of the section, we consider this issue.


\subsection{Incorporating Distillation Loss}
We consider the point-wise ranking loss for binary relevance labels for performing distillation, we also tried the pair-wise loss and will discuss their pros and cons later.
Similar to Eqn~(\ref{eqn:rankloss_point}), we formalize distillation loss as:
\begin{equation}\label{eqn:rdloss_general}
\begin{aligned}
\mathcal{L}^{D}(\boldsymbol{\pi}_{1..K}, \boldsymbol{\hat{y}})=&-\sum_{r=1}^{K} w_r \cdot \text{log}(P(rel=1|\hat{y}_{\pi_r}))\\
=&-\sum_{r=1}^{K} w_r \cdot \text{log}(\sigma(\hat{y}_{\pi_r})),
\end{aligned}
\end{equation}
where $\sigma(\cdot)$ is the \emph{sigmoid} function and $w_r$ is the weight to be discussed later. There are several differences compared to Eqn~(\ref{eqn:rankloss_point}).
First, in Eqn~(\ref{eqn:rdloss_general}), we treat the top-$K$ ranked documents from the teacher model as positive instances and there is no negative instance. Recall that KD causes the student model to output a higher probability for the label ``tiger'' when the ground-truth label is ``cat'' because their features captured by the teacher model are correlated.
Along this line, we want the student model to rank higher for teacher's top-$K$ ranked documents.
As we mentioned above, for the given query,
 besides the ground-truth positive documents $\boldsymbol{y}^{+}$,
teacher's top-$K$ ranked unlabeled documents are also strongly correlated to this query. These correlations are captured by the well-trained powerful teacher model in the latent space when using latent factor model or neural networks.

However, as $K$ increases, the relevance of the top-$K$ ranked unlabeld documents
becomes weaker. Following the work of learning from noise labels~\cite{natarajan2013learning}, 
we use a weighted sum over the loss on documents from $\boldsymbol{\pi}_{1..K}$ with weight $w_r$ on each position $r$ from $1$ to $K$. There are two straightforward choices for $w_r$: $w_r = 1/r$ puts more emphasis on the top positions, whereas  $w_r = 1/K$ weights each position equally. Such weightings are heuristic and
pre-determined, may not be flexible enough to deal with general cases. 
Instead, we introduce two flexible weighting schemes, which were shown to be superior
in our experimental studies.

\subsubsection{Weighting by Position Importance}
In this weighting scheme, we assume that the teacher predicted unlabeled documents at top positions are more correlated to the query and are more likely to the positive ground-truth documents, therefore, this weight $w^{a}$ should be inversely proportional to the rank:
\begin{equation}\label{eqn:position_weight1}
w^{a}_r \propto r^{-1}\quad \textrm{and} \quad r \in [1,K],
\end{equation}
where $r$ is the rank range from $1$ to $K$. As pointed out above,
this scheme pre-determines the weight.
Rendle \emph{et al}~\cite{rendle2014improving} proposed an empirical weight for sampling a single position from a top-$K$ ranking, following a geometric distribution:
\begin{equation}\label{eqn:position_weight2}
w^{a}_r = \rho(1-\rho)^{r}\quad \textrm{and} \quad \rho \in (0,1).
\end{equation}
Following their work, we use a parametrized geometric distribution for weighting the position importance:
\begin{equation}\label{eqn:position_weight3}
w^{a}_r \propto e^{-r/\lambda} \quad \textrm{and} \quad \lambda \in \mathbb{R}^{+},
\end{equation}
where $\lambda$ is the hyperparameter that controls the sharpness of the distribution, and is searched through the validation set. When $\lambda$ is small, this scheme puts more emphasis on top positions, and when $\lambda$ is large enough, the distribution becomes the uniform distribution. This parametrization is easy to implement and configurable to each kind of situation.


\begin{algorithm}[t]
\caption{Estimate Student's Ranking for $\pi_r$}
\begin{algorithmic}
\REQUIRE Student Model $M_S(q,d;\theta_S)$, unlabeled document set $\mathcal{\bar{O}}$ for a given query $q$ and the hyper-parameter $\epsilon$\\
\STATE $\hat{y}_{\pi_r} \leftarrow M_S(q,\pi_r;\theta_S)$
\STATE Initialize $n=0$
\FOR{$t=1,2,...\epsilon$}
\STATE Sample a document $d$ from $\mathcal{\bar{O}}$ without replacement
\STATE $\hat{y}_{d} \leftarrow M_S(q,d;\theta_S)$
\IF{$\hat{y}_{d} > \hat{y}_{\pi_r}$}
\STATE $n \leftarrow n+1$
\ENDIF
\ENDFOR
\STATE $\hat{r}_{\pi_r}\leftarrow \floor[\big]{ \frac{n\times (|\mathcal{\bar{O}}|-1)}{\epsilon} }+1$
\RETURN $\hat{r}_{\pi_r}$
\end{algorithmic}
\label{alg:cal_rank}
\end{algorithm}

\subsubsection{Weighting by Ranking Discrepancy}
The weighting by position importance is static, meaning that the weight at the same position is fixed during training process. Our second scheme is dynamic that considers the discrepancy between the student-predicted rank and the teacher-predicted rank for a given unlabeled document, and uses it as another weight ${w}^{b}$. This weighting scheme allows the training to gradually concentrate on the documents in teacher's top-$K$ ranking that are not well-predicted by the student. The details are as follows.



For the $r$-th document $\pi_r~(r\in[1,K])$ in teacher model's top-$K$ ranking, the teacher-predicted ranking~(\emph{i.e.,} $r$) is known for us. But we know only the student predicted relevant score $\hat{y}_{\pi_r}$ instead of its rank without computing relevance scores for all documents. To get the student predicted rank for this document, we apply Weston~\emph{et al}~\cite{weston2010large}'s sequential sampling, and do it in a parallel manner~\cite{hsieh2017collaborative}. 
As described in Algorithm~\ref{alg:cal_rank}, for the $r$-th document $\pi_r$, if we want to know its rank in a list of $N$ documents without computing the scores for all documents, we can randomly sample $\epsilon \in [1, N-1]$ documents in this list and estimate the relative rank by $n/\epsilon$, where $n$ is the number of documents whose (student) scores are greater than $\hat{y}_{\pi_r}$. Then the estimated rank in the whole list is $\hat{r}_{\pi_r}=\floor[\big]{\frac{n \times (N-1)}{\epsilon}}+1$. When $\epsilon$ goes larger, the estimated rank is more close to the actual rank.

\begin{figure}[t!]  
\centering
\includegraphics[scale=0.17]{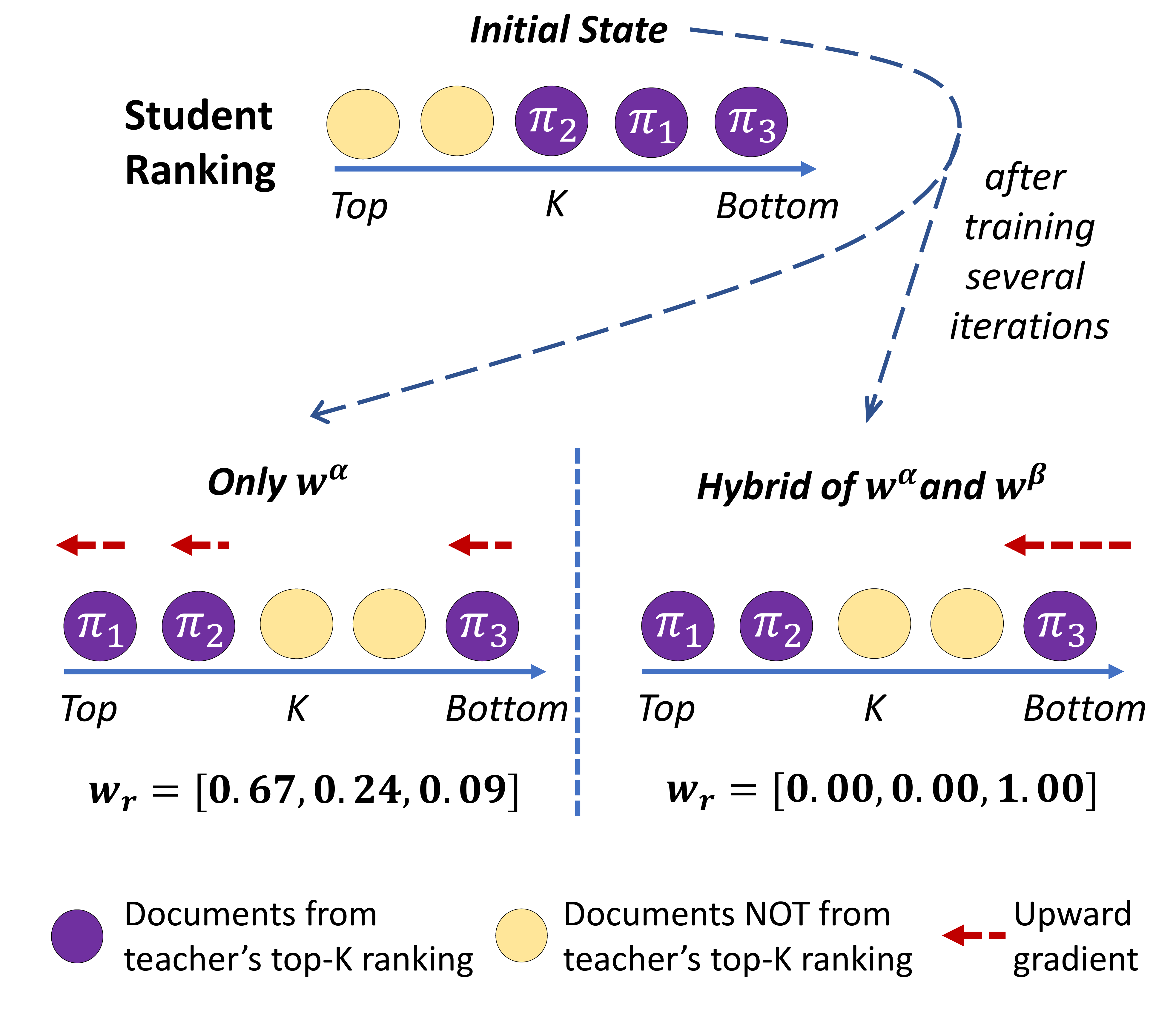}
\vspace{-0.0cm}
\caption{ An illustration of hybrid weighting scheme. We use $K=3$ in this example. }\label{fig:hybrid_weight}
\end{figure}

After getting the estimated student's rank $\hat{r}_{\pi_r}$ for the $r$-th document $\pi_r$ in teacher's top-$K$ ranking, the discrepancy between $r$ and $\hat{r}$ is computed by
\begin{equation}\label{eqn:dynamic_weight}
w^{b}_r = \textit{tanh}(\textrm{max}(\mu \cdot (\hat{r}_{\pi_r} - r), 0)),
\end{equation}
where $\textit{tanh}(\cdot)$ is a rescaled \emph{logistic} function $\textit{tanh}(x)=2\sigma(2x)-1$ that rescale the output range to $[0,1]$ when $x>0$. The hyper-parameter $\mu \in \mathbb{R}^{+}$ is used to control the sharpness of the \emph{tanh} function. Eqn (\ref{eqn:dynamic_weight}) gives a dynamic weight: when the student predicted-rank of a document is close to its teacher, we think this document has been well-predicted and impose little loss on it~(\emph{i.e.,} $w^{b}_r \approx 0$); the rest concentrates on the documents~(\emph{i.e.,} $w^{b}_r \approx 1$) whose student predicted-rank is far from the teacher's rank. Note that the ranking discrepancy weight ${w}^{b}$ is computed for each document in $\pi_{1..K}$ during training. So in practice, we choose $\epsilon \ll |\hat{\mathcal{O}}|$ for training efficiency. While extra computation used to compute relevance scores for sampled $\epsilon$ documents, we still boost the whole offline training process. Because the dynamic weight allows the training to focus on the erroneous parts in the distillation loss.


\begin{table*}[t!]
\center
\caption{Statistics of the data sets}\label{tb:dataset}
\vspace{-0.0cm}
\setlength{\tabcolsep}{13pt}
\begin{tabular}{lccccc}
\toprule
\multirow{2}{4em}{\textbf{Datasets}} &
 \multirow{2}{3em}{\textbf{\#users}} & \multirow{2}{3em}{\textbf{\#items}} &
 \textbf{avg. actions} &
 \textbf{$(u, \mathcal{S}^{(u,t)})$} & \multirow{2}{3em}{\textbf{Sparsity}}\\
 & & & \textbf{per user} & \textbf{pairs} &\\
\midrule
Gowalla & 13.1k & 14.0k & 40.74 & 367.6k & 99.71\%\\
\midrule
Foursquare & 10.1k & 23.4k & 30.16 & 198.9k & 99.87\%\\
\bottomrule
\end{tabular}
\end{table*}
\subsubsection{Hybrid Weighting Scheme}
The hybrid weighting combines the weight $w^{a}$ by position importance, and the weight $w^{b}$ by ranking discrepancy: $w_r = {(w^{a}_r \cdot w^{b}_r)  / (\sum_{i=1}^K w^{a}_i \cdot w^{b}_i)}$. Figure~\ref{fig:hybrid_weight} illustrates the advantages of hybrid weighting over weighting only by position importance.
Our experiments show that this hybrid weighting gives better results in general.
In the actual implementation, since the estimated student ranking of $\hat{r}_{\pi_{r}}$ 
is not accurate during the first few iterations, we use only $w^{a}$ during the first $m$ iterations to warm up the model, and then use the hybrid weighting to make training focus on the erroneous parts in distillation loss. $m$ should be determined via the validation set. In our experiments, $m$ is usually set to more than half of the total training iterations.

\subsection{Discussion}
Under the paradigm of ranking distillation, for a certain query $q$, besides the labeled documents, we use a top-$K$ ranking for unlabeled documents generated by a well-trained teacher ranking model $M_T$ as extra information to guide the training of the student ranking model $M_S$ with less parameters. During the student model training, we use a weighted point-wise ranking loss as the distillation loss and propose two types of flexible weighting schemes, \emph{i.e.}, ${w}^{a}$ and ${w}^{b}$ and propose an effective way to fusion them. 
For the hyper-parameters~($\alpha,\lambda,\mu,\epsilon, K, m$),
they are dataset-dependent and are determined for each data set through the validation set. 
Two key factors for the success of ranking distillation are: 
(1) larger models are
capable to capture the complex interaction patterns between queries and documents, thus, their predicted unlabeled documents at top positions are also strongly correlated with the given query and 
(2) student models with less parameters can learn from the extra-provided helpful information~(top-$K$ unlabeled documents in teacher's ranking) and boost their performances.

We also tried to use a pair-wise distillation loss when learning from teacher's top-$K$ ranking. Specifically, we use Eqn~(\ref{eqn:rankloss_pair}) for the distillation loss by taking the partial order in teacher's top-$K$ ranking as objective.
However, the results were disappointing. We found that if we use pair-wise distillation loss to place much focus on the partial order within teacher's ranking, it will produce both upward and downward gradients, making the training unstable and sometimes even fail to converge. However, our weighted point-wise distillation loss that only contains upward gradients doesn't suffer from this issue.

\section{Experimental Studies}\label{sec:exper}
 We evaluate the performance of \emph{ranking distillation} on two real-world data sets. 
The source code and processed data sets are publicly available online\footnote{https://github.com/graytowne/rank\_distill}.

\subsection{Experimental Setup}
\noindent{\textbf{Task Description}.}
We use recommendation as our task for evaluating the performance of ranking distillation. In this problem, we have a set of users $\mathcal{U} = \{u_1, u_2,\cdots,u_{|\mathcal{U}|}\}$ and a universe of items  $\mathcal{I} = \{i_1, i_2,\cdots,i_{|\mathcal{I}|}\}$. For recommendation without context information, we can cache the recommendation list for each user
\footnote{We suppose the recommendation model doesn't change immediately whenever new observed data come, which is common in real-world cases.}.
However, for context-aware recommendation, we have to re-compute the recommendation list each time a user comes with a new context, so the online inference efficiency becomes important. The following \emph{sequential recommendation} is
one case of context-aware recommendation.  
Given a users $u$ with her/his history sequence~(\emph{i.e.,} past $L$ interacted items) at time $t$, $\mathcal{S}^{(u,t)}=(\mathcal{S}^{u}_{t-1},..,\mathcal{S}^{u}_{t-L})$, where $\mathcal{S}^u_i \in \mathcal{I}$, the goal is to retrieve a list of items for this user that meets her/his future needs. In IR's terms, the query is the user profile $(u,\mathcal{S}^{(u,t)})$ at time $t$, and the document is the item.
Note that whenever the user has a new behavior~(\emph{e.g.,} watch a video/listen to a music), 
we have to re-compute the recommendation list as her/his context changes. We also wish to point out that, in general, ranking distillation can be applied to other learning to rank tasks, not just to recommendation. 


\noindent{\textbf{Datasets}.}
We choose two real-world data sets in this work, as they contain numerous sequential signals and thus suitable for sequential recommendation~\cite{tang2018caser}. Their statistics are described in Table~\ref{tb:dataset}.
Gowalla\footnote{https://snap.stanford.edu/data/loc-gowalla.html} was constructed by~\cite{cho2011friendship} and Foursquare was obtained from \cite{yuan2014graph}. These data sets contain sequences of implicit feedbacks through user-venue check-ins.
During the offline training phase, for a user $u$, we extract every 5 successive items~($L=5$) from her sequence as $\mathcal{S}^{(u,t)}$, and the immediately next item as the ground-truth.
Following \cite{tang2018caser,yuan2014graph},
we hold the first 70\% of actions in each user's sequence as the \emph{training set} and use the next 10\% of actions as the \emph{validation set} to search the optimal hyperparameter settings for all models. The remaining 20\% actions in each user's sequence are used as the \emph{test set} for evaluating a model's performance.

\noindent{\textbf{Evaluation Metrics}.}
As in ~\cite{pang2017deeprank,xiong2017ena,dehghani2017neural,pang2016tmi}, three different evaluation metrics used are Precision@$n$ (Prec@$n$), nDCG@$n$, and Mean Average Precision (MAP). We set $n\in \{3, 5, 10\}$, as recommendations
are top positions of rank lists are more important. To measure the online inference efficiency, we count the number of parameters in each model and report the wall time for making a recommendation list to every user based on her/his last 5 actions in the training data set. While training models efficiently is also important, training is done offline 
before the recommendation phase starts, and our focus is the online inference efficiency where the user 
is waiting for the responses from the system.  

\begin{table*}[t!]
\caption{Performance comparison. (1) The performance of the models with ranking distillation, 
Fossil-RD and Caser-RD, always has statistically significant improvements over the student-only models, Fossil-S and Caser-S.
(2) The performance of the models with ranking distillation, Fossil-RD and Caser-RD, has no significant degradation
from that of the teacher models, Fossil-T and Caser-T.
We use the one-tail t-test with significance level at 0.05. }
\vspace{-0.1cm}
\setlength{\tabcolsep}{0.20cm}
\centering
\label{tb:perform}
\begin{tabular}{l|lllllll}

\multicolumn{8}{c}{\textbf{Gowalla}}\\
\toprule
\textbf{Model} & \textbf{Prec@3} & \textbf{Prec@5} & \textbf{Prec@10} & \textbf{nDCG@3} & \textbf{nDCG@5} & \textbf{nDCG@10} & \textbf{MAP}\\
\midrule
\midrule
\emph{Fossil-T} & 0.1299 & 0.1062 & 0.0791 & 0.1429 & 0.1270 & 0.1140 & 0.0866\\
\emph{Fossil-RD} & 0.1355 & 0.1096 & 0.0808 & 0.1490 & 0.1314 & 0.1172 & 0.0874\\
\emph{Fossil-S} & 0.1217 & 0.0995 & 0.0739 & 0.1335 & 0.1185 & 0.1060 & 0.0792\\

\midrule
\emph{Caser-T} & 0.1408 & 0.1149 & 0.0856 & 0.1546 & 0.1376 & 0.1251 & 0.0958\\
\emph{Caser-RD} & 0.1458 & 0.1183 & 0.0878 & 0.1603 & 0.1423 & 0.1283 & 0.0969\\
\emph{Caser-S} & 0.1333 & 0.1094 & 0.0818 & 0.1456 & 0.1304 & 0.1188 & 0.0919\\

\midrule
\emph{POP} & 0.0341 & 0.0362 & 0.0281 & 0.0517 & 0.0386 & 0.0344 & 0.0229\\
\emph{ItemCF} & 0.0686 & 0.0610 & 0.0503 & 0.0717 & 0.0675 & 0.0640 & 0.0622\\
\emph{BPR} & 0.1204 & 0.0983 & 0.0726 & 0.1301 & 0.1155 & 0.1037 & 0.0767\\
\bottomrule

\multicolumn{8}{c}{}\\
\multicolumn{8}{c}{\textbf{Foursquare}}\\

\toprule
\textbf{Model} & \textbf{Prec@3} & \textbf{Prec@5} & \textbf{Prec@10} & \textbf{nDCG@3} & \textbf{nDCG@5} & \textbf{nDCG@10} & \textbf{MAP}\\
\midrule
\midrule
\emph{Fossil-T} & 0.0859 & 0.0630 & 0.0420 & 0.1182 & 0.1085 & 0.1011 & 0.0891 \\
\emph{Fossil-RD} & 0.0877 & 0.0648 & 0.0430 & 0.1203 & 0.1102 & 0.1023 & 0.0901 \\
\emph{Fossil-S} & 0.0766 & 0.0556 & 0.0355 & 0.1079 & 0.0985 & 0.0911 & 0.0780 \\

\midrule
\emph{Caser-T} & 0.0860 & 0.0650 & 0.0438 & 0.1182 & 0.1105 & 0.1041 &0.0941\\
\emph{Caser-RD} & 0.0923 & 0.0671 & 0.0444 & 0.1261 & 0.1155 & 0.1076 & 0.0952\\
\emph{Caser-S} & 0.0830 & 0.0621 & 0.0413 & 0.1134 & 0.1051 & 0.0986 & 0.0874\\

\midrule
\emph{POP} & 0.0702 & 0.0477 & 0.0304 & 0.0845 & 0.0760 & 0.0706 & 0.0636\\
\emph{ItemCF} & 0.0248 & 0.0221 & 0.0187 & 0.0282 & 0.0270 & 0.0260 & 0.0304\\
\emph{BPR} & 0.0744 & 0.0543 & 0.0348 & 0.0949 & 0.0871 & 0.0807 & 0.0719\\
\bottomrule
\end{tabular}
\end{table*}

\noindent{\textbf{Teacher/Student Models}}.
We apply the proposed ranking distillation to two sequential recommendation models that have been shown to have strong performances:
\begin{itemize}
\item[$\bullet$] \textbf{Fossil}.
Factorized Sequential Prediction with Item Similarity ModeLs~(Fossil)~\cite{HeMcA16b} models sequential patterns and user preferences by fusing a similarity model with latent factor model. It uses a pair-wise ranking loss.

\item[$\bullet$] \textbf{Caser}.
ConvolutionAl Sequence Embedding Recommendation model~(Caser)~\cite{tang2018caser} incorporates the Convolutional Neural Network and latent factor model to learn sequential patterns as well as user preferences. It uses a point-wise ranking loss.
\end{itemize}

To apply ranking distillation, we adopt as many parameters as possible for the teacher model to achieve a good performance on each data set. These well-trained teacher models are denoted by \textbf{Fossil-T} and \textbf{Caser-T}.
We then use these models to teach smaller student models denoted by \textbf{Fossil-RD} and \textbf{Caser-RD} by minimizing the ranking distillation loss in~Eqn~(\ref{eqn:rd1}). The model sizes of the student models are gradually increased until the models reach a comparable performance to their teachers. \textbf{Fossil-S} and \textbf{Caser-S} denote
the student models trained with only ranking loss, \emph{i.e.}, without the help from the teacher. Note that the increasing in model sizes is achieved by using larger dimensions for embeddings, without any changes to the model structure.

\subsection{Overall Results}\label{sec:exper_compare}
The results of each method are summarized in Table~\ref{tb:perform}.
We also included three non-sequential recommendation baselines: the popularity (in all users' sequences) based item recommendation~(\textbf{POP}), the item based Collaborative Filtering\footnote{We use Jaccard similarity measure and set the number of nearest neighbor to 20.}~(\textbf{ItemCF})~\cite{sarwar2001item}, and the Bayesian personalized ranking~(\textbf{BPR})~\cite{rendle2009bpr}. Clearly, the performance of these non-sequential baselines is worse than that of the sequential recommenders, \emph{i.e.,} Fossil and Caser.

\begin{table}[t!]
\center
\caption{Model compactness and online inference efficiency. Time (seconds) indicates the wall time used for generating a recommendation list for every user. Ratio is the student model's parameter size relative to the teacher model's parameter size. }\label{tb:rd_time}
\vspace{-0.1cm}
\setlength{\tabcolsep}{0.2cm}
\begin{tabular}{l|l|cccc}
\toprule
\multirow{2}{4em}{\textbf{Datasets}} &
 \multirow{2}{3em}{\textbf{Model}} & \textbf{Time} & \textbf{Time} & \multirow{2}{3.2em}{\textbf{\#Params}} & \multirow{2}{2.7em}{\textbf{Ratio}}\\
& & (\textbf{CPU}) & (\textbf{GPU}) & &\\

\midrule
\midrule
\multirow{4}{3em}{Gowalla} & \emph{Fossil-T} & 9.32s & 3.72s & 1.48M & 100\%\\
 & \emph{Fossil-RD} & 4.99s & 2.11s & 0.64M & 43.2\%\\
 & \emph{Caser-T} & 38.58s & 4.52s & 5.58M & 100\%\\
 & \emph{Caser-RD} & 18.63s & 2.99s & 2.79M & 50.0\%\\
\midrule
\multirow{4}{3em}{Foursquare} & \emph{Fossil-T} & 6.35s & 2.47s & 1.01M & 100\%\\
 & \emph{Fossil-RD} & 3.86s & 2.01s & 0.54M & 53.5\%\\
 & \emph{Caser-T} & 23.89s & 2.95s & 4.06M & 100\%\\
 & \emph{Caser-RD} & 11.65s & 1.96s & 1.64M & 40.4\%\\
\bottomrule
\end{tabular}
\end{table}

\begin{figure*}[t!]  

        \centering
        \begin{subfigure}[b]{0.5\textwidth}
                \includegraphics[width=\textwidth]{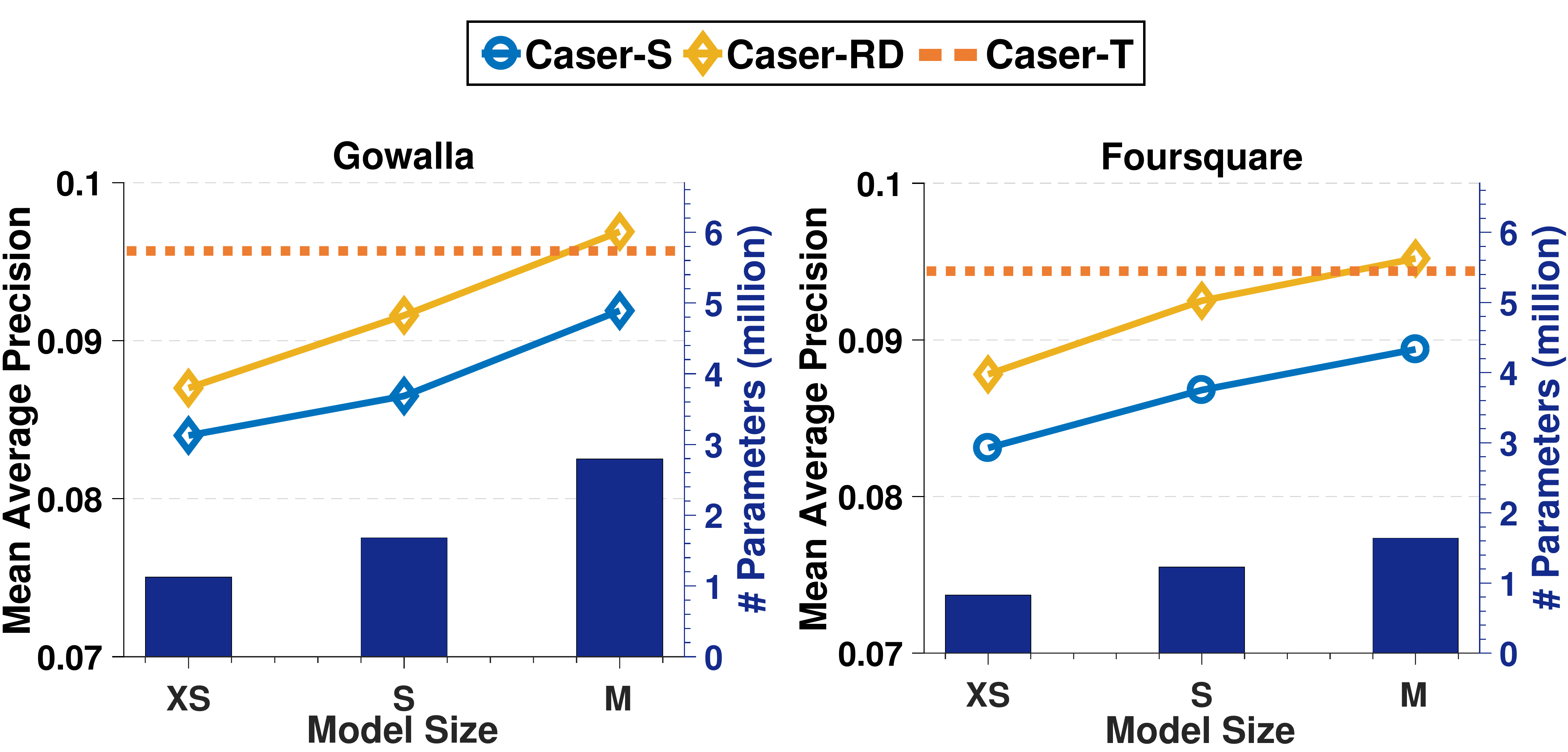}
                \caption{MAP vs. model size}
                \label{fig:map_d}
        \end{subfigure}%
        ~
        \hspace{0.05in}
        \begin{subfigure}[b]{0.48\textwidth}
                \includegraphics[width=\textwidth]{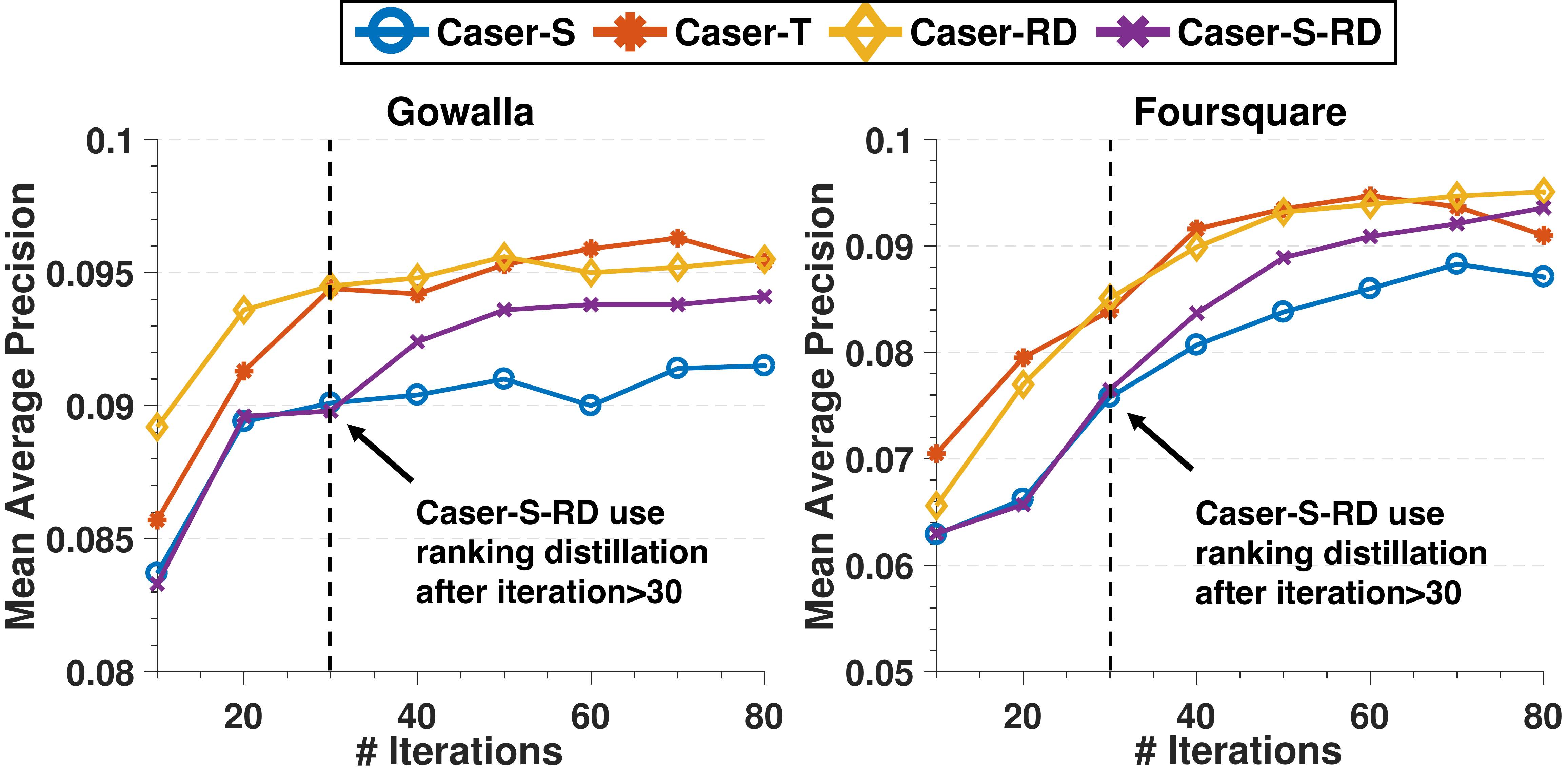}
                \caption{MAP vs. the number of iterations for model training}
                \label{fig:map_iter}
        \end{subfigure}%

        \vspace{-0.0cm}
        \caption{
        Mean average precision vs. (a) model size and (b) the choice of distillation loss.}\label{fig:map_d_iter}

\end{figure*}

The teacher models, \emph{i.e.}, Fossil-T and Caser-T, have a better performance than the student-only models, \emph{i.e.}, Fossil-S and Caser-S, indicating that a larger model size provides more flexibility to fit the complex data with more predictive power.
The effectiveness of ranking distillation is manifested by the significantly better performance of Fossil-RD and Caser-RD compared to Fossil-S and Caser-S, and by the similar performance of Fossil-RD and Caser-RD compared to Fossil-T and Caser-T.
In other words, thanks to the knowledge transfer of ranking distillation, we are able to learn a student model
that has fewer parameters but similar performance 
as the teacher model. Surprisingly, student models with ranking distillation often have even better performance than their teachers. This finding is consistent with \cite{kim2016sequence} and we will explain possible reasons in Section \ref{sec:exper_source}. 

The online inference efficiency is measured by the model size~( number of model parameters) and is shown in Table~\ref{tb:rd_time}. Note that Fossil-S and Caser-S have the same model size as Fossil-RD and Caser-RD. All inferences were implemented using PyTorch with CUDA from GTX1070 GPU and Intel i7-6700K CPU. Fossil-RD and Caser-RD  
nearly half down the model size compared to their teacher models,  Fossil-T and Caser-T. This reduction in model size
is translated into a similar reduction in online inference time. In many practical applications, 
the data set is much larger than the data sets considered here in terms of the numbers of users and items;
for example, Youtube could have 30 million active users per day and 1.3 billion of items\footnote{https://fortunelords.com/youtube-statistics}. For such large data sets, online inference
could be more time-consuming and the reduction in model size has more privileges. Also, 
for models that are much more complicated than Fossil and Caser, the reduction in model size 
could yield a larger reduction in online inference time than reported here.

In conclusion, the findings in Table~\ref{tb:perform} and ~\ref{tb:rd_time} together confirm 
that ranking distillation helps generate compact models with no or little compromise on effectiveness, 
and these advantages are independent of the choices of models.

\begin{figure}[t!]  

        \centering
        \begin{subfigure}[b]{0.24\textwidth}
        \includegraphics[width=\textwidth]{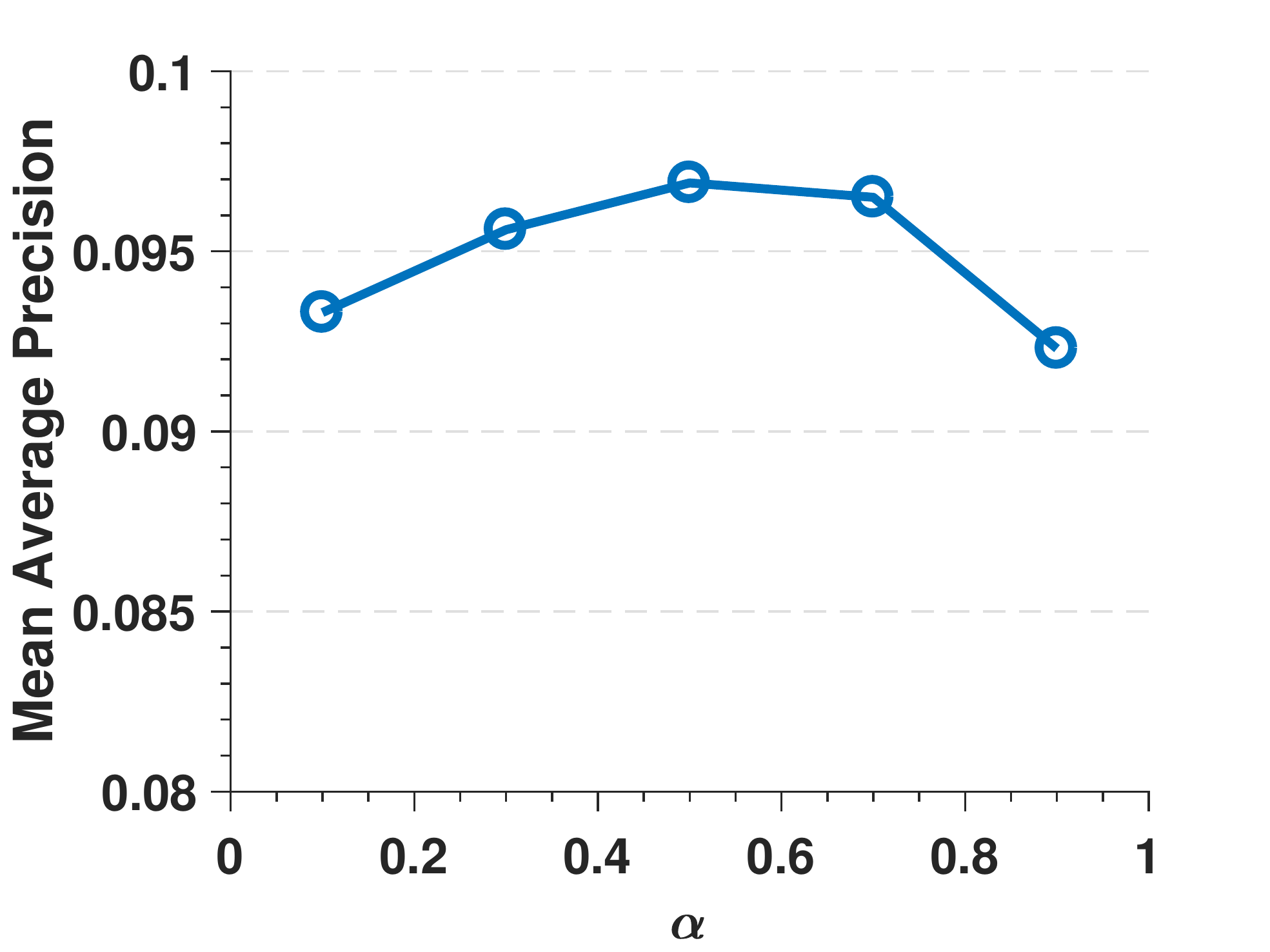}
        \caption{Gowalla}
        \end{subfigure}%
        ~
        \begin{subfigure}[b]{0.24\textwidth}                            \includegraphics[width=\textwidth]{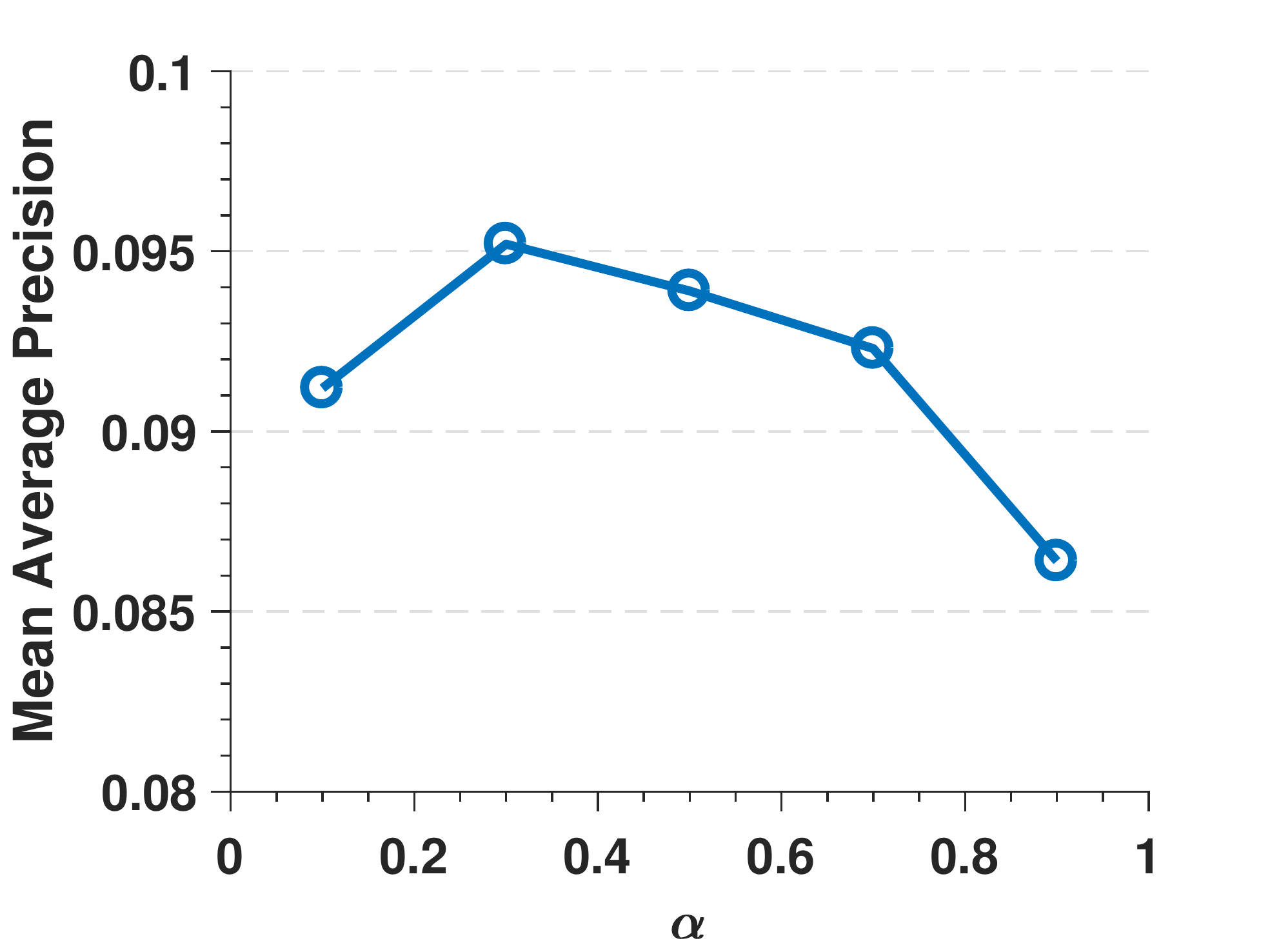}
                \caption{Foursquare}
        \end{subfigure}
        \caption{MAP vs. balancing parameter $\alpha$}\label{fig:map_alpha}

\end{figure}

\subsection{Effects of Model Size and Distillation Loss}\label{sec:exper_source}

In this experiment, we study the impact of model size on the student model's performance~(\emph{i.e.,} MAP).
We consider only Caser because the results for Fossil are similar.
Figure~\ref{fig:map_d} shows the results. 
Caser-S and Caser-RD perform better when the model size goes up, but there is always a gap. Caser-RD reaches a similar performance to its teacher with the medium model size, which is about 50\% of the teacher model size. 

Figure~\ref{fig:map_iter} shows the impact of ranking distillation 
on the student model's MAP iteration by iteration. We compare four models: Case-S, Caser-T, Caser-RD, and 
Caser-S-RD. The last model minimizes ranking loss during the first 30 iterations and adds distillation 
loss after that. Caser-RD outperforms Caser-S all the time. Caser-S reaches its limit after 50 iterations on Gowalla and 70 iterations on Foursquare. Caser-S-RD performs similarly to Caser-S during the first 30 iterations, but catches up with the Caser-RD at the end, indicating the impressive effectiveness of ranking distillation. 
Caser-T performs well at first but tends to get overfitted after about 60 iterations due to its large model size and the sparse recommendation data sets. In contrast, Caser-RD and Caser-S-RD, which have smaller model sizes, are more robust to overfitting issue, though their training is partially supervised by the teacher. This finding reveals another 
advantage of ranking distillation. 

Figure~\ref{fig:map_alpha} shows the MAP for various balancing parameter $\alpha$ to explore models' performance when balancing ranking loss and distillation loss. For Gowalla data, the best performance is achieved when $\alpha$ is around 0.5. But for Foursquare data, the best performance is achieved when $\alpha$ is around 0.3, indicating too much concentrate on distillation loss leads to a bad performance. On both data sets, either discarding ranking loss or discarding distillation loss gives poor results.

\subsection{Effects of Weighting Schemes}
Table~\ref{tb:rd_weight} shows the effects of the proposed weighting schemes in our ranking distillation. For the weight $w_r$ for $r$-th document in the teacher's top-$K$ ranking, we used the equal weight ($w_r=1/K$) as the baseline and considered the weighting by position importance ($w_r=w^{a}_r$), the weighting by ranking discrepancy ($w_r=w^{b}_r$), and the hybrid weighting ($w_r \propto w^{a}_r \cdot w^{b}_r$). The equal weight performs the worst. The position importance weighting is much better, suggesting that within the teacher's top-$K$ ranking, documents at top positions
are more related to the positive ground truth. The ranking discrepancy weighting only doesn't give impressive results, 
but when used with the position importance weighting, the hybrid weighting yields the best results on both data sets.

\begin{table}[t!]
\center
\caption{Performance of Caser-RD with different choices of weighting scheme on two data sets.}\label{tb:rd_weight}
\vspace{-0.0cm}
\setlength{\tabcolsep}{0.2cm}
\begin{tabular}{l|c|cccc}
\toprule
\textbf{Datasets} & {\centering \textbf{Weighting}} & \textbf{P@10} & \textbf{nDCG@10}& \textbf{MAP}\\
\midrule
\midrule
\multirow{4}{4em}{Gowalla} & $w_r=1/K$ & 0.0843  & 0.1198 & 0.0925\\
 & $w_r=w^{a}_r$ & 0.0850 & 0.1230 & 0.0945\\
 & $w_r=w^{b}_r$ & 0.0851 & 0.1227 & 0.0937\\
 & hybrid & \textbf{0.0878} & \textbf{0.1283} & \textbf{0.0969}\\
\midrule
\multirow{4}{4em}{Foursquare} & $w_r=1/K$ & 0.0424 & 0.1046 & 0.0914\\
 & $w_r = w^{a}_r$ & 0.0423 & 0.1052 & 0.0929 \\
 & $w_r = w^{b}_r$ & 0.0429 & 0.1035 & 0.0912\\
 & hybrid & \textbf{0.0444} & \textbf{0.1076} & \textbf{0.0952}\\
\bottomrule
\end{tabular}
\end{table} 

\section{Related Work}\label{sec:related}
In this section, we compared our works with several related research areas.

\noindent{\textbf{Knowledge Distillation}}
Knowledge distillation has been used in image recognition~\cite{hinton2015distilling, ba2014deep, romero2014fitnets} and neural machine translation~\cite{kim2016sequence} as a way to generate compact models. 
As pointed out in Introduction, it is not straightforward to apply KD 
to ranking models and new issues must be addressed. In the context of ranking problems, 
the most relevant work is~\cite{chen2017darkrank}, which uses knowledge distillation for image retrieval.
This method applies the sampling technique to rank a sample of the image from all data each time. In general, training on a 
sample works if the sample shares similar patterns with the rest of data through some
content information, such as image contents in the case of \cite{chen2017darkrank}. But this technique is not applicable to  training a recommender model when items and users are represented by IDs with no content information, 
as in the case of collaborative filtering. In this case, the recommender model training cannot be easily generalize to all users and items.

\noindent{\textbf{Semi-Supervised Learning}}
Another related research area is semi-supervised learning~\cite{zhu2005semi, chapelle2009semi}. Unlike the teacher-student model learning paradigm in knowledge distillation and in our work, semi-supervised learning usually trains a single model and utilizes weak-labeled or unlabeled data as well as the labeled data to gain a better performance. Several works in information retrieval followed this direction, using weak-labeled or unlabeled data to construct test collections~\cite{asadi2011pseudo}, to provide extra features~\cite{diaz2016learning} and labels~\cite{dehghani2017neural} for ranking model training.
The basic idea of ranking distillation and semi-supervised learning is similar as they both utilize unlabeled data while with different purpose.

\noindent{\textbf{Transfer Learning for Recommender System}}
Transfer learning has been widely used in the field of recommender systems~\cite{chen2016learning,fernandez2012cross}.
These methods mainly focus on how to transfer knowledge (\emph{e.g.}, user rating patterns) 
from a source domain~(\emph{e.g.}, movies) to a target domain~(\emph{e.g.,} musics) for improving the recommendation performance. If we consider the student as a target model and the teacher as a source model, our teacher-student learning  
can be seen as a special transfer learning. However, unlike transfer learning, our teacher-student learning does not 
require two domains because the teacher and student models are learned from the same domain. 
Having a compact student model to enhance online inference efficiency is another purpose of our teacher-student learning. 
\section{Conclusion}\label{sec:conclu}
The proposed ranking distillation enables generating compact ranking models for better online inference efficiency without scarifying the ranking performance. The idea is training a teacher model with more parameters to teach a student model with fewer parameters to rank unlabeled documents. While the student model is compact, its 
training benefits from the extra supervision of the teacher, in addition to the usual ground truth from the training data,
making the student model comparable with the teacher model in the ranking performance. This paper focused
on several key issues of ranking distillation, \emph{i.e.}, the problem formulation, the representation of teacher's supervision, and the balance between the trust on the training data and the trust on the teacher, and presented our solutions.
The evaluation on real data sets supported our claims. 

\section*{Acknowledgement}
The work of the second author is partially supported by a Discovery Grant from Natural Sciences and Engineering Research Council of Canada.



\bibliographystyle{ACM-Reference-Format}
\balance
{\small
\bibliography{bib/abbrv_kdd15}


\begin{thebibliography}{43}


\ifx \showCODEN    \undefined \def \showCODEN     #1{\unskip}     \fi
\ifx \showDOI      \undefined \def \showDOI       #1{#1}\fi
\ifx \showISBNx    \undefined \def \showISBNx     #1{\unskip}     \fi
\ifx \showISBNxiii \undefined \def \showISBNxiii  #1{\unskip}     \fi
\ifx \showISSN     \undefined \def \showISSN      #1{\unskip}     \fi
\ifx \showLCCN     \undefined \def \showLCCN      #1{\unskip}     \fi
\ifx \shownote     \undefined \def \shownote      #1{#1}          \fi
\ifx \showarticletitle \undefined \def \showarticletitle #1{#1}   \fi
\ifx \showURL      \undefined \def \showURL       {\relax}        \fi
\providecommand\bibfield[2]{#2}
\providecommand\bibinfo[2]{#2}
\providecommand\natexlab[1]{#1}
\providecommand\showeprint[2][]{arXiv:#2}

\bibitem[\protect\citeauthoryear{Anil, Pereyra, Passos, Orm{\'{a}}ndi, E.~Dahl,
  and E.~Hinton}{Anil et~al\mbox{.}}{2018}]%
        {Anil2018Large}
\bibfield{author}{\bibinfo{person}{Rohan Anil}, \bibinfo{person}{Gabriel
  Pereyra}, \bibinfo{person}{Alexandre Passos}, \bibinfo{person}{Robert
  Orm{\'{a}}ndi}, \bibinfo{person}{George E.~Dahl}, {and}
  \bibinfo{person}{Geoffrey E.~Hinton}.} \bibinfo{year}{2018}\natexlab{}.
\newblock \showarticletitle{Large scale distributed neural network training
  through online distillation}.
\newblock \bibinfo{journal}{\emph{arXiv preprint arXiv:1804.03235}}
  (\bibinfo{year}{2018}).
\newblock


\bibitem[\protect\citeauthoryear{Asadi, Metzler, Elsayed, and Lin}{Asadi
  et~al\mbox{.}}{2011}]%
        {asadi2011pseudo}
\bibfield{author}{\bibinfo{person}{Nima Asadi}, \bibinfo{person}{Donald
  Metzler}, \bibinfo{person}{Tamer Elsayed}, {and} \bibinfo{person}{Jimmy
  Lin}.} \bibinfo{year}{2011}\natexlab{}.
\newblock \showarticletitle{Pseudo test collections for learning web search
  ranking functions}. In \bibinfo{booktitle}{\emph{International Conference on
  Research and development in Information Retrieval}}. ACM,
  \bibinfo{pages}{1073--1082}.
\newblock


\bibitem[\protect\citeauthoryear{Ba and Caruana}{Ba and Caruana}{2014}]%
        {ba2014deep}
\bibfield{author}{\bibinfo{person}{Jimmy Ba} {and} \bibinfo{person}{Rich
  Caruana}.} \bibinfo{year}{2014}\natexlab{}.
\newblock \showarticletitle{Do deep nets really need to be deep?}. In
  \bibinfo{booktitle}{\emph{Advances in neural information processing
  systems}}. \bibinfo{pages}{2654--2662}.
\newblock


\bibitem[\protect\citeauthoryear{Chapelle, Scholkopf, and Zien}{Chapelle
  et~al\mbox{.}}{2009}]%
        {chapelle2009semi}
\bibfield{author}{\bibinfo{person}{Olivier Chapelle}, \bibinfo{person}{Bernhard
  Scholkopf}, {and} \bibinfo{person}{Alexander Zien}.}
  \bibinfo{year}{2009}\natexlab{}.
\newblock \showarticletitle{Semi-supervised learning (chapelle, o. et al.,
  eds.; 2006)[book reviews]}.
\newblock \bibinfo{journal}{\emph{IEEE Transactions on Neural Networks}}
  \bibinfo{volume}{20}, \bibinfo{number}{3} (\bibinfo{year}{2009}),
  \bibinfo{pages}{542--542}.
\newblock


\bibitem[\protect\citeauthoryear{Chen, Qin, Zhang, and Xu}{Chen
  et~al\mbox{.}}{2016}]%
        {chen2016learning}
\bibfield{author}{\bibinfo{person}{Xu Chen}, \bibinfo{person}{Zheng Qin},
  \bibinfo{person}{Yongfeng Zhang}, {and} \bibinfo{person}{Tao Xu}.}
  \bibinfo{year}{2016}\natexlab{}.
\newblock \showarticletitle{Learning to rank features for recommendation over
  multiple categories}. In \bibinfo{booktitle}{\emph{International ACM SIGIR
  conference on Research and Development in Information Retrieval}}.
  \bibinfo{pages}{305--314}.
\newblock


\bibitem[\protect\citeauthoryear{Chen, Wang, and Zhang}{Chen
  et~al\mbox{.}}{2017}]%
        {chen2017darkrank}
\bibfield{author}{\bibinfo{person}{Yuntao Chen}, \bibinfo{person}{Naiyan Wang},
  {and} \bibinfo{person}{Zhaoxiang Zhang}.} \bibinfo{year}{2017}\natexlab{}.
\newblock \showarticletitle{DarkRank: Accelerating Deep Metric Learning via
  Cross Sample Similarities Transfer}.
\newblock \bibinfo{journal}{\emph{arXiv preprint arXiv:1707.01220}}
  (\bibinfo{year}{2017}).
\newblock


\bibitem[\protect\citeauthoryear{Cho, Myers, and Leskovec}{Cho
  et~al\mbox{.}}{2011}]%
        {cho2011friendship}
\bibfield{author}{\bibinfo{person}{Eunjoon Cho}, \bibinfo{person}{Seth~A
  Myers}, {and} \bibinfo{person}{Jure Leskovec}.}
  \bibinfo{year}{2011}\natexlab{}.
\newblock \showarticletitle{Friendship and mobility: user movement in
  location-based social networks}. In \bibinfo{booktitle}{\emph{International
  Conference on Knowledge Discovery and Data Mining}}. ACM,
  \bibinfo{pages}{1082--1090}.
\newblock


\bibitem[\protect\citeauthoryear{Choromanska, Henaff, Mathieu, Arous, and
  LeCun}{Choromanska et~al\mbox{.}}{2015}]%
        {choromanska2015loss}
\bibfield{author}{\bibinfo{person}{Anna Choromanska}, \bibinfo{person}{Mikael
  Henaff}, \bibinfo{person}{Michael Mathieu}, \bibinfo{person}{G{\'e}rard~Ben
  Arous}, {and} \bibinfo{person}{Yann LeCun}.} \bibinfo{year}{2015}\natexlab{}.
\newblock \showarticletitle{The loss surfaces of multilayer networks}. In
  \bibinfo{booktitle}{\emph{Artificial Intelligence and Statistics}}.
  \bibinfo{pages}{192--204}.
\newblock


\bibitem[\protect\citeauthoryear{Covington, Adams, and Sargin}{Covington
  et~al\mbox{.}}{2016}]%
        {covington2016deep}
\bibfield{author}{\bibinfo{person}{Paul Covington}, \bibinfo{person}{Jay
  Adams}, {and} \bibinfo{person}{Emre Sargin}.}
  \bibinfo{year}{2016}\natexlab{}.
\newblock \showarticletitle{Deep neural networks for youtube recommendations}.
  In \bibinfo{booktitle}{\emph{ACM Conference on Recommender systems}}.
  \bibinfo{pages}{191--198}.
\newblock


\bibitem[\protect\citeauthoryear{Dehghani, Zamani, Severyn, Kamps, and
  Croft}{Dehghani et~al\mbox{.}}{2017}]%
        {dehghani2017neural}
\bibfield{author}{\bibinfo{person}{Mostafa Dehghani}, \bibinfo{person}{Hamed
  Zamani}, \bibinfo{person}{Aliaksei Severyn}, \bibinfo{person}{Jaap Kamps},
  {and} \bibinfo{person}{W~Bruce Croft}.} \bibinfo{year}{2017}\natexlab{}.
\newblock \showarticletitle{Neural Ranking Models with Weak Supervision}.
\newblock \bibinfo{journal}{\emph{arXiv preprint arXiv:1704.08803}}
  (\bibinfo{year}{2017}).
\newblock


\bibitem[\protect\citeauthoryear{Diaz}{Diaz}{2016}]%
        {diaz2016learning}
\bibfield{author}{\bibinfo{person}{Fernando Diaz}.}
  \bibinfo{year}{2016}\natexlab{}.
\newblock \showarticletitle{Learning to Rank with Labeled Features}. In
  \bibinfo{booktitle}{\emph{International Conference on the Theory of
  Information Retrieval}}. ACM, \bibinfo{pages}{41--44}.
\newblock


\bibitem[\protect\citeauthoryear{Fern{\'a}ndez-Tob{\'\i}as, Cantador,
  Kaminskas, and Ricci}{Fern{\'a}ndez-Tob{\'\i}as et~al\mbox{.}}{2012}]%
        {fernandez2012cross}
\bibfield{author}{\bibinfo{person}{Ignacio Fern{\'a}ndez-Tob{\'\i}as},
  \bibinfo{person}{Iv{\'a}n Cantador}, \bibinfo{person}{Marius Kaminskas},
  {and} \bibinfo{person}{Francesco Ricci}.} \bibinfo{year}{2012}\natexlab{}.
\newblock \showarticletitle{Cross-domain recommender systems: A survey of the
  state of the art}. In \bibinfo{booktitle}{\emph{Spanish Conference on
  Information Retrieval}}. sn, \bibinfo{pages}{24}.
\newblock


\bibitem[\protect\citeauthoryear{He and McAuley}{He and McAuley}{2016}]%
        {HeMcA16b}
\bibfield{author}{\bibinfo{person}{Ruining He} {and} \bibinfo{person}{Julian
  McAuley}.} \bibinfo{year}{2016}\natexlab{}.
\newblock \showarticletitle{Fusing Similarity Models with Markov Chains for
  Sparse Sequential Recommendation}. In \bibinfo{booktitle}{\emph{International
  Conference on Data Mining}}. \bibinfo{publisher}{IEEE}.
\newblock


\bibitem[\protect\citeauthoryear{He, Liao, Zhang, Nie, Hu, and Chua}{He
  et~al\mbox{.}}{2017}]%
        {he2017neural}
\bibfield{author}{\bibinfo{person}{Xiangnan He}, \bibinfo{person}{Lizi Liao},
  \bibinfo{person}{Hanwang Zhang}, \bibinfo{person}{Liqiang Nie},
  \bibinfo{person}{Xia Hu}, {and} \bibinfo{person}{Tat-Seng Chua}.}
  \bibinfo{year}{2017}\natexlab{}.
\newblock \showarticletitle{Neural collaborative filtering}. In
  \bibinfo{booktitle}{\emph{International Conference on World Wide Web}}. ACM,
  \bibinfo{pages}{173--182}.
\newblock


\bibitem[\protect\citeauthoryear{Hinton, Vinyals, and Dean}{Hinton
  et~al\mbox{.}}{2015}]%
        {hinton2015distilling}
\bibfield{author}{\bibinfo{person}{Geoffrey Hinton}, \bibinfo{person}{Oriol
  Vinyals}, {and} \bibinfo{person}{Jeff Dean}.}
  \bibinfo{year}{2015}\natexlab{}.
\newblock \showarticletitle{Distilling the knowledge in a neural network}.
\newblock \bibinfo{journal}{\emph{arXiv preprint arXiv:1503.02531}}
  (\bibinfo{year}{2015}).
\newblock


\bibitem[\protect\citeauthoryear{Hsieh, Yang, Cui, Lin, Belongie, and
  Estrin}{Hsieh et~al\mbox{.}}{2017}]%
        {hsieh2017collaborative}
\bibfield{author}{\bibinfo{person}{Cheng-Kang Hsieh}, \bibinfo{person}{Longqi
  Yang}, \bibinfo{person}{Yin Cui}, \bibinfo{person}{Tsung-Yi Lin},
  \bibinfo{person}{Serge Belongie}, {and} \bibinfo{person}{Deborah Estrin}.}
  \bibinfo{year}{2017}\natexlab{}.
\newblock \showarticletitle{Collaborative metric learning}. In
  \bibinfo{booktitle}{\emph{International Conference on World Wide Web}}.
  \bibinfo{pages}{193--201}.
\newblock


\bibitem[\protect\citeauthoryear{Karpathy, Toderici, Shetty, Leung, Sukthankar,
  and Fei-Fei}{Karpathy et~al\mbox{.}}{2014}]%
        {karpathy2014large}
\bibfield{author}{\bibinfo{person}{Andrej Karpathy}, \bibinfo{person}{George
  Toderici}, \bibinfo{person}{Sanketh Shetty}, \bibinfo{person}{Thomas Leung},
  \bibinfo{person}{Rahul Sukthankar}, {and} \bibinfo{person}{Li Fei-Fei}.}
  \bibinfo{year}{2014}\natexlab{}.
\newblock \showarticletitle{Large-scale video classification with convolutional
  neural networks}. In \bibinfo{booktitle}{\emph{IEEE conference on Computer
  Vision and Pattern Recognition}}. \bibinfo{pages}{1725--1732}.
\newblock


\bibitem[\protect\citeauthoryear{Kawaguchi}{Kawaguchi}{2016}]%
        {kawaguchi2016deep}
\bibfield{author}{\bibinfo{person}{Kenji Kawaguchi}.}
  \bibinfo{year}{2016}\natexlab{}.
\newblock \showarticletitle{Deep learning without poor local minima}. In
  \bibinfo{booktitle}{\emph{Advances in Neural Information Processing
  Systems}}. \bibinfo{pages}{586--594}.
\newblock


\bibitem[\protect\citeauthoryear{Kim}{Kim}{2014}]%
        {yoon14convolution}
\bibfield{author}{\bibinfo{person}{Yoon Kim}.} \bibinfo{year}{2014}\natexlab{}.
\newblock \showarticletitle{Convolutional Neural Networks for Sentence
  Classification}. In \bibinfo{booktitle}{\emph{Conference on Empirical Methods
  on Natural Language Processing}}. ACL, \bibinfo{pages}{1756--1751}.
\newblock


\bibitem[\protect\citeauthoryear{Kim and Rush}{Kim and Rush}{2016}]%
        {kim2016sequence}
\bibfield{author}{\bibinfo{person}{Yoon Kim} {and} \bibinfo{person}{Alexander~M
  Rush}.} \bibinfo{year}{2016}\natexlab{}.
\newblock \showarticletitle{Sequence-level knowledge distillation}.
\newblock \bibinfo{journal}{\emph{arXiv preprint arXiv:1606.07947}}
  (\bibinfo{year}{2016}).
\newblock


\bibitem[\protect\citeauthoryear{Koren, Bell, and Volinsky}{Koren
  et~al\mbox{.}}{2009}]%
        {koren2009matrix}
\bibfield{author}{\bibinfo{person}{Yehuda Koren}, \bibinfo{person}{Robert
  Bell}, {and} \bibinfo{person}{Chris Volinsky}.}
  \bibinfo{year}{2009}\natexlab{}.
\newblock \showarticletitle{Matrix factorization techniques for recommender
  systems}.
\newblock \bibinfo{journal}{\emph{Computer}} \bibinfo{volume}{42},
  \bibinfo{number}{8} (\bibinfo{year}{2009}).
\newblock


\bibitem[\protect\citeauthoryear{Krizhevsky, Sutskever, and Hinton}{Krizhevsky
  et~al\mbox{.}}{2012}]%
        {krizhevsky2012imagenet}
\bibfield{author}{\bibinfo{person}{Alex Krizhevsky}, \bibinfo{person}{Ilya
  Sutskever}, {and} \bibinfo{person}{Geoffrey~E Hinton}.}
  \bibinfo{year}{2012}\natexlab{}.
\newblock \showarticletitle{Imagenet classification with deep convolutional
  neural networks}. In \bibinfo{booktitle}{\emph{Advances in Neural Information
  Processing Systems}}. \bibinfo{pages}{1097--1105}.
\newblock


\bibitem[\protect\citeauthoryear{Li, Chan, Yiu, and Mamoulis}{Li
  et~al\mbox{.}}{2017}]%
        {li2017fexipro}
\bibfield{author}{\bibinfo{person}{Hui Li}, \bibinfo{person}{Tsz~Nam Chan},
  \bibinfo{person}{Man~Lung Yiu}, {and} \bibinfo{person}{Nikos Mamoulis}.}
  \bibinfo{year}{2017}\natexlab{}.
\newblock \showarticletitle{FEXIPRO: Fast and Exact Inner Product Retrieval in
  Recommender Systems}. In \bibinfo{booktitle}{\emph{International Conference
  on Management of Data}}. ACM, \bibinfo{pages}{835--850}.
\newblock


\bibitem[\protect\citeauthoryear{Liu, Rogers, Shiau, Kislyuk, Ma, Zhong, Liu,
  and Jing}{Liu et~al\mbox{.}}{2017}]%
        {liu2017related}
\bibfield{author}{\bibinfo{person}{David~C Liu}, \bibinfo{person}{Stephanie
  Rogers}, \bibinfo{person}{Raymond Shiau}, \bibinfo{person}{Dmitry Kislyuk},
  \bibinfo{person}{Kevin~C Ma}, \bibinfo{person}{Zhigang Zhong},
  \bibinfo{person}{Jenny Liu}, {and} \bibinfo{person}{Yushi Jing}.}
  \bibinfo{year}{2017}\natexlab{}.
\newblock \showarticletitle{Related pins at pinterest: The evolution of a
  real-world recommender system}. In \bibinfo{booktitle}{\emph{International
  Conference on World Wide Web}}. \bibinfo{pages}{583--592}.
\newblock


\bibitem[\protect\citeauthoryear{Mikolov, Karafi{\'a}t, Burget, Cernock{\`y},
  and Khudanpur}{Mikolov et~al\mbox{.}}{2010}]%
        {mikolov2010recurrent}
\bibfield{author}{\bibinfo{person}{Tomas Mikolov}, \bibinfo{person}{Martin
  Karafi{\'a}t}, \bibinfo{person}{Lukas Burget}, \bibinfo{person}{Jan
  Cernock{\`y}}, {and} \bibinfo{person}{Sanjeev Khudanpur}.}
  \bibinfo{year}{2010}\natexlab{}.
\newblock \showarticletitle{Recurrent neural network based language model.}. In
  \bibinfo{booktitle}{\emph{Interspeech}}.
\newblock


\bibitem[\protect\citeauthoryear{Natarajan, Dhillon, Ravikumar, and
  Tewari}{Natarajan et~al\mbox{.}}{2013}]%
        {natarajan2013learning}
\bibfield{author}{\bibinfo{person}{Nagarajan Natarajan},
  \bibinfo{person}{Inderjit~S Dhillon}, \bibinfo{person}{Pradeep~K Ravikumar},
  {and} \bibinfo{person}{Ambuj Tewari}.} \bibinfo{year}{2013}\natexlab{}.
\newblock \showarticletitle{Learning with noisy labels}. In
  \bibinfo{booktitle}{\emph{Advances in neural information processing
  systems}}. \bibinfo{pages}{1196--1204}.
\newblock


\bibitem[\protect\citeauthoryear{Pang, Lan, Guo, Xu, Wan, and Cheng}{Pang
  et~al\mbox{.}}{2016}]%
        {pang2016tmi}
\bibfield{author}{\bibinfo{person}{Liang Pang}, \bibinfo{person}{Yanyan Lan},
  \bibinfo{person}{Jiafeng Guo}, \bibinfo{person}{Jun Xu},
  \bibinfo{person}{Shengxian Wan}, {and} \bibinfo{person}{Xueqi Cheng}.}
  \bibinfo{year}{2016}\natexlab{}.
\newblock \showarticletitle{Text Matching As Image Recognition}. In
  \bibinfo{booktitle}{\emph{AAAI Conference on Artificial Intelligence}}.
  \bibinfo{publisher}{AAAI Press}, \bibinfo{pages}{2793--2799}.
\newblock


\bibitem[\protect\citeauthoryear{Pang, Lan, Guo, Xu, Xu, and Cheng}{Pang
  et~al\mbox{.}}{2017}]%
        {pang2017deeprank}
\bibfield{author}{\bibinfo{person}{Liang Pang}, \bibinfo{person}{Yanyan Lan},
  \bibinfo{person}{Jiafeng Guo}, \bibinfo{person}{Jun Xu},
  \bibinfo{person}{Jingfang Xu}, {and} \bibinfo{person}{Xueqi Cheng}.}
  \bibinfo{year}{2017}\natexlab{}.
\newblock \showarticletitle{DeepRank: A New Deep Architecture for Relevance
  Ranking in Information Retrieval}.
\newblock \bibinfo{journal}{\emph{arXiv preprint arXiv:1710.05649}}
  (\bibinfo{year}{2017}).
\newblock


\bibitem[\protect\citeauthoryear{Rendle and Freudenthaler}{Rendle and
  Freudenthaler}{2014}]%
        {rendle2014improving}
\bibfield{author}{\bibinfo{person}{Steffen Rendle} {and}
  \bibinfo{person}{Christoph Freudenthaler}.} \bibinfo{year}{2014}\natexlab{}.
\newblock \showarticletitle{Improving pairwise learning for item recommendation
  from implicit feedback}. In \bibinfo{booktitle}{\emph{International
  Conference on Web Search and Data Mining}}. ACM, \bibinfo{pages}{273--282}.
\newblock


\bibitem[\protect\citeauthoryear{Rendle, Freudenthaler, Gantner, and
  Schmidt-Thieme}{Rendle et~al\mbox{.}}{2009}]%
        {rendle2009bpr}
\bibfield{author}{\bibinfo{person}{Steffen Rendle}, \bibinfo{person}{Christoph
  Freudenthaler}, \bibinfo{person}{Zeno Gantner}, {and} \bibinfo{person}{Lars
  Schmidt-Thieme}.} \bibinfo{year}{2009}\natexlab{}.
\newblock \showarticletitle{BPR: Bayesian personalized ranking from implicit
  feedback}. In \bibinfo{booktitle}{\emph{Conference on Uncertainty in
  Artificial Intelligence}}. AUAI Press, \bibinfo{pages}{452--461}.
\newblock


\bibitem[\protect\citeauthoryear{Romero, Ballas, Kahou, Chassang, Gatta, and
  Bengio}{Romero et~al\mbox{.}}{2014}]%
        {romero2014fitnets}
\bibfield{author}{\bibinfo{person}{Adriana Romero}, \bibinfo{person}{Nicolas
  Ballas}, \bibinfo{person}{Samira~Ebrahimi Kahou}, \bibinfo{person}{Antoine
  Chassang}, \bibinfo{person}{Carlo Gatta}, {and} \bibinfo{person}{Yoshua
  Bengio}.} \bibinfo{year}{2014}\natexlab{}.
\newblock \showarticletitle{Fitnets: Hints for thin deep nets}.
\newblock \bibinfo{journal}{\emph{arXiv preprint arXiv:1412.6550}}
  (\bibinfo{year}{2014}).
\newblock


\bibitem[\protect\citeauthoryear{Sarwar, Karypis, Konstan, and Riedl}{Sarwar
  et~al\mbox{.}}{2001}]%
        {sarwar2001item}
\bibfield{author}{\bibinfo{person}{Badrul Sarwar}, \bibinfo{person}{George
  Karypis}, \bibinfo{person}{Joseph Konstan}, {and} \bibinfo{person}{John
  Riedl}.} \bibinfo{year}{2001}\natexlab{}.
\newblock \showarticletitle{Item-based collaborative filtering recommendation
  algorithms}. In \bibinfo{booktitle}{\emph{International Conference on World
  Wide Web}}. ACM, \bibinfo{pages}{285--295}.
\newblock


\bibitem[\protect\citeauthoryear{Tang and Wang}{Tang and Wang}{2018}]%
        {tang2018caser}
\bibfield{author}{\bibinfo{person}{Jiaxi Tang} {and} \bibinfo{person}{Ke
  Wang}.} \bibinfo{year}{2018}\natexlab{}.
\newblock \showarticletitle{Personalized Top-N Sequential Recommendation via
  Convolutional Sequence Embedding}. In \bibinfo{booktitle}{\emph{ACM
  International Conference on Web Search and Data Mining}}.
\newblock


\bibitem[\protect\citeauthoryear{Teflioudi, Gemulla, and Mykytiuk}{Teflioudi
  et~al\mbox{.}}{2015}]%
        {teflioudi2015lemp}
\bibfield{author}{\bibinfo{person}{Christina Teflioudi},
  \bibinfo{person}{Rainer Gemulla}, {and} \bibinfo{person}{Olga Mykytiuk}.}
  \bibinfo{year}{2015}\natexlab{}.
\newblock \showarticletitle{Lemp: Fast retrieval of large entries in a matrix
  product}. In \bibinfo{booktitle}{\emph{International Conference on Management
  of Data}}. ACM, \bibinfo{pages}{107--122}.
\newblock


\bibitem[\protect\citeauthoryear{Wang, Wang, and Yeung}{Wang
  et~al\mbox{.}}{2015}]%
        {wang2015collaborative}
\bibfield{author}{\bibinfo{person}{Hao Wang}, \bibinfo{person}{Naiyan Wang},
  {and} \bibinfo{person}{Dit-Yan Yeung}.} \bibinfo{year}{2015}\natexlab{}.
\newblock \showarticletitle{Collaborative deep learning for recommender
  systems}. In \bibinfo{booktitle}{\emph{International Conference on Knowledge
  Discovery and Data Mining}}. ACM, \bibinfo{pages}{1235--1244}.
\newblock


\bibitem[\protect\citeauthoryear{Weston, Bengio, and Usunier}{Weston
  et~al\mbox{.}}{2010}]%
        {weston2010large}
\bibfield{author}{\bibinfo{person}{Jason Weston}, \bibinfo{person}{Samy
  Bengio}, {and} \bibinfo{person}{Nicolas Usunier}.}
  \bibinfo{year}{2010}\natexlab{}.
\newblock \showarticletitle{Large scale image annotation: learning to rank with
  joint word-image embeddings}.
\newblock \bibinfo{journal}{\emph{Machine learning}} \bibinfo{volume}{81},
  \bibinfo{number}{1} (\bibinfo{year}{2010}), \bibinfo{pages}{21--35}.
\newblock


\bibitem[\protect\citeauthoryear{Xiong, Dai, Callan, Liu, and Power}{Xiong
  et~al\mbox{.}}{2017}]%
        {xiong2017ena}
\bibfield{author}{\bibinfo{person}{Chenyan Xiong}, \bibinfo{person}{Zhuyun
  Dai}, \bibinfo{person}{Jamie Callan}, \bibinfo{person}{Zhiyuan Liu}, {and}
  \bibinfo{person}{Russell Power}.} \bibinfo{year}{2017}\natexlab{}.
\newblock \showarticletitle{End-to-End Neural Ad-hoc Ranking with Kernel
  Pooling}. In \bibinfo{booktitle}{\emph{International ACM SIGIR conference on
  Research and Development in Information Retrieval}}. \bibinfo{pages}{55--64}.
\newblock


\bibitem[\protect\citeauthoryear{Yuan, Cong, and Sun}{Yuan
  et~al\mbox{.}}{2014}]%
        {yuan2014graph}
\bibfield{author}{\bibinfo{person}{Quan Yuan}, \bibinfo{person}{Gao Cong},
  {and} \bibinfo{person}{Aixin Sun}.} \bibinfo{year}{2014}\natexlab{}.
\newblock \showarticletitle{Graph-based point-of-interest recommendation with
  geographical and temporal influences}. In
  \bibinfo{booktitle}{\emph{International Conference on Information and
  Knowledge Management}}. ACM, \bibinfo{pages}{659--668}.
\newblock


\bibitem[\protect\citeauthoryear{Zhang, Shen, Liu, He, Luan, and Chua}{Zhang
  et~al\mbox{.}}{2016}]%
        {zhang2016discrete}
\bibfield{author}{\bibinfo{person}{Hanwang Zhang}, \bibinfo{person}{Fumin
  Shen}, \bibinfo{person}{Wei Liu}, \bibinfo{person}{Xiangnan He},
  \bibinfo{person}{Huanbo Luan}, {and} \bibinfo{person}{Tat-Seng Chua}.}
  \bibinfo{year}{2016}\natexlab{}.
\newblock \showarticletitle{Discrete collaborative filtering}. In
  \bibinfo{booktitle}{\emph{International Conference on Research and
  Development in Information Retrieval}}. ACM, \bibinfo{pages}{325--334}.
\newblock


\bibitem[\protect\citeauthoryear{Zhang, Lian, and Yang}{Zhang
  et~al\mbox{.}}{2017}]%
        {zhang2017discrete}
\bibfield{author}{\bibinfo{person}{Yan Zhang}, \bibinfo{person}{Defu Lian},
  {and} \bibinfo{person}{Guowu Yang}.} \bibinfo{year}{2017}\natexlab{}.
\newblock \showarticletitle{Discrete Personalized Ranking for Fast
  Collaborative Filtering from Implicit Feedback.}. In
  \bibinfo{booktitle}{\emph{AAAI Conference on Artificial Intelligence}}. AAAI
  Press, \bibinfo{pages}{1669--1675}.
\newblock


\bibitem[\protect\citeauthoryear{Zhang, Wang, Ruan, and Si}{Zhang
  et~al\mbox{.}}{2014}]%
        {zhang2014preference}
\bibfield{author}{\bibinfo{person}{Zhiwei Zhang}, \bibinfo{person}{Qifan Wang},
  \bibinfo{person}{Lingyun Ruan}, {and} \bibinfo{person}{Luo Si}.}
  \bibinfo{year}{2014}\natexlab{}.
\newblock \showarticletitle{Preference preserving hashing for efficient
  recommendation}. In \bibinfo{booktitle}{\emph{International Conference on
  Research and Development in Information Retrieval}}. ACM,
  \bibinfo{pages}{183--192}.
\newblock


\bibitem[\protect\citeauthoryear{Zhou and Zha}{Zhou and Zha}{2012}]%
        {zhou2012learning}
\bibfield{author}{\bibinfo{person}{Ke Zhou} {and} \bibinfo{person}{Hongyuan
  Zha}.} \bibinfo{year}{2012}\natexlab{}.
\newblock \showarticletitle{Learning binary codes for collaborative filtering}.
  In \bibinfo{booktitle}{\emph{International Conference on Knowledge Discovery
  and Data Mining}}. ACM, \bibinfo{pages}{498--506}.
\newblock


\bibitem[\protect\citeauthoryear{Zhu}{Zhu}{2005}]%
        {zhu2005semi}
\bibfield{author}{\bibinfo{person}{Xiaojin Zhu}.}
  \bibinfo{year}{2005}\natexlab{}.
\newblock \showarticletitle{Semi-supervised learning literature survey}.
\newblock  (\bibinfo{year}{2005}).
\newblock


\end{thebibliography}
}

\end{document}